\UseRawInputEncoding
\documentclass{article}

\usepackage{PRIMEarxiv}

\usepackage{subcaption}
\usepackage{caption}
\usepackage{moresize}
\usepackage{amsmath}
\usepackage[utf8]{inputenc} 
\usepackage[T1]{fontenc}    
\usepackage{hyperref}       
\usepackage{url}            
\usepackage{booktabs}       
\usepackage{amsfonts}       
\usepackage{nicefrac}       
\usepackage[numbers]{natbib}
\usepackage{microtype}      
\usepackage{xspace} 
\usepackage{lipsum}
\usepackage{fancyhdr}       
\usepackage{graphicx}       
\usepackage{listings}
\graphicspath{{media/}}     

\newcommand{\datasetname}{MCTED\xspace}
\title{\datasetname: A Machine-Learning-Ready Dataset for Digital Elevation Model Generation From Mars Imagery}

\author{
  Rafał Osadnik\\
  European Space Agency (ESA)\\
  European Space Astronomy Centre (ESAC) \\
  Villaneuva de la Cañada\\
  Madrid, Spain\\
  \texttt{rafal.osadnik@ext.esa.int} \\
  \And
  Pablo Gómez \\
  European Space Agency (ESA)\\
  European Space Astronomy Centre (ESAC) \\
  Villaneuva de la Cañada\\
  Madrid, Spain\\
  \texttt{pablo.gomez@esa.int} \\
   \And
  Eleni Bohacek \\
  UK Research and Innovation - Innovate UK \\
  Swindon, United Kingdom\\
  \texttt{eleni.bohacek@iuk.ukri.org} \\
  \And
  Rickbir Bahia \\
  UK Space Agency (UKSA) \\
  Harwell, United Kingdom\\
  \texttt{rickbir.bahia@ukspaceagency.gov.uk} \\
}

\begin{document}

\maketitle

\begin{abstract}
This work presents a new dataset for the Martian digital elevation model prediction task, ready for machine learning applications called \datasetname. The dataset has been generated using a comprehensive pipeline designed to process high-resolution Mars orthoimage and DEM pairs from \citeauthor{day2023mars}, yielding a dataset consisting of 80,898 data samples. The source images are data gathered by the Mars Reconnaissance Orbiter using the CTX instrument, providing a very diverse and comprehensive coverage of the Martian surface. Given the complexity of the processing pipelines used in large-scale DEMs, there are often artefacts and missing data points in the original data, for which we developed tools to solve or mitigate their impact. We divide the processed samples into training and validation splits, ensuring samples in both splits cover no mutual areas to avoid data leakage. Every sample in the dataset is represented by the optical image patch, DEM patch, and two mask patches, indicating values that were originally missing or were altered by us. This allows future users of the dataset to handle altered elevation regions as they please. We provide statistical insights of the generated dataset, including the spatial distribution of samples, the distributions of elevation values, slopes and more.  Finally, we train a small U-Net architecture on the \datasetname dataset and compare its performance to a monocular depth estimation foundation model, DepthAnythingV2, on the task of elevation prediction. We find that even a very small architecture trained on this dataset specifically, beats a zero-shot performance of a depth estimation foundation model like DepthAnythingV2. We make the dataset and code used for its generation completely open source in public repositories.
\\\\
Source code: \href{https://github.com/ESA/MCTED}{https://github.com/ESA/MCTED}

\end{abstract}

\section{Introduction}

Digital Elevation Models (DEMs) provide three-dimensional insights into outcrops and landscapes, enabling scientific inquiries beyond the capabilities of conventional imaging. For Mars, DEMs are integral for numerous scientific and explorative endeavours, ranging from landing site mapping for rovers (e.g., \cite{fawdon_2024}), landers, and future human exploration, to modelling of ancient hydrological processes (e.g., \cite{Bahia_2022}). While global DEMs of Mars are accessible at coarse resolutions ($\sim$460 meters per pixel) \cite{mola}, the higher-resolution DEMs necessary for detailed scientific investigations and landing site analyses remain limited in coverage.  Furthermore, the traditional methodology for generating high-resolution DEMs via the NASA Ames Stereo Pipeline is highly resource-intensive. It necessitates stereo pair images: two images of the same area captured from different angles. If stereo pairs are not initially available for a desired location, obtaining an additional image requires a request, potentially leading to significant delays in DEM production. Moreover, multiple processing steps are needed to convert stereo pair images into a DEM \cite{Ohman_2013}, ultimately resulting in a DEM with a resolution lower than that of the original images. DEMs made from a single image could solve a number of these drawbacks.

At present, the Mars Reconnaissance Orbiter and its Context Camera (CTX) have acquired more than 100,000 separate panchromatic images that capture nearly the entire surface (>99 \%) of Mars at $\sim$5–6 m/pixel \cite{dickson} with over 60 \% of the planet being imaged more than once, allowing for the generation of stereo pairs. However, to date, the number of CTX DEMs is limited, covering just a small part of the Martian surface \cite{day2023mars}. Additionally, $\sim$4 \% of the surface has been imaged by the HiRISE camera, at $\sim$0.25 to 0.5 m per pixel, with 12 \% of these having DEMs produced via the traditional stereo pair method \cite{mcewen2024}.

When a 3-D scene is projected and recorded as a 2-D plane during imaging, the distance between the scene objects or depth information is lost. Depth estimation or reconstruction aims to solve the inverse and ill-posed problem of recovering the 3-D scene structure from a 2-D image. This has been an intensely researched area for decades. Approaches such as stereo and photogrammetric reconstruction use multiple images of the scene by detecting corresponding points, calculating disparity and relative depth, and then absolute depth by incorporating calibration information. Monocular depth estimation aims to reconstruct the 3-D scene using a single image. Using traditional computer vision approaches, it is only possible by applying assumptions and limitations; however, recent deep learning approaches show great promise for monocular depth estimation without \textit{a priori} scene knowledge \cite{Arampatzakis_2024, bhat2023zoedepth, Yin_2023_ICCV, depthv2}.

Machine-learning and deep-learning methods, by extension, are based on processing large amounts of data to extract patterns, learn and exploit data relationships to generate results that fit a desired objective, usually by minimising a loss function. The more data a machine-learning model has seen during training, the better it usually is, assuming the data is of high quality and represents the problem the model is trying to solve accurately. There are not a lot of publicly available datasets, meant for depth estimation from the orbital perspective, suitable for DEM generation using single perspective images. In this work, we make the following contributions:
\begin{itemize}
    \item We provide an extensive analysis of the \citeauthor{day2023mars} repository, identifying and mitigating data quality challenges in DEMs generated by the NASA Ames Stereo Pipeline.
    \item We create and provide \datasetname, a cleaned, chunked and curated dataset of 80,898 image and DEM pairs with supplementary masks, suited for training machine learning DEM generation models. We make it publicly available in a HuggingFace repository, under the following address: \href{https://huggingface.co/datasets/ESA-Datalabs/MCTED}{https://huggingface.co/datasets/ESA-Datalabs/MCTED}.
    \item We demonstrate the usefulness of \datasetname by training a small U-Net architecture able to outperform state-of-the-art MDE foundation models in the DEM generation task, showing they are not suited for this task by default.
    \item We make the source code for dataset cleaning, generation, model training and evaluation open source under the following address: \href{https://github.com/ESA/MCTED}{https://github.com/ESA/MCTED}.
\end{itemize}

\section{Related Work}

The development and application of digital elevation models (DEMs) for Mars have significantly advanced our understanding of its topography, geology \cite{mars_valleys, Bahia_2022}, and potential habitability sites \cite{mars_habibability}. Numerous DEMs have been generated over the past two decades \cite{day2023mars, mola, blended_mola} using various instruments onboard different spacecraft missions. These datasets vary in resolution, spatial coverage, and methods of acquisition, each contributing unique insights but also presenting challenges for machine learning applications.

In this section, we first explore the existing literature on Mars elevation maps to provide a comprehensive overview of current DEMs available. We then transition to an examination of the Context Camera (CTX) instrument aboard the Mars Reconnaissance Orbiter (MRO), highlighting its capabilities and contributions to Martian surface studies. Following that, we explore recent advancements in monocular depth estimation techniques and their application in generating elevation models from satellite imagery. Lastly, we discuss existing datasets suitable for training machine learning models for monocular depth estimation, noting their limitations and biases.

\subsection{Mars Elevation Maps}

Many missions orbiting Mars regularly capture images of its surface. Notable examples include the Mars Reconnaissance Orbiter (NASA), Mars Odyssey (NASA), Mars Express (ESA), and the Trace Gas Orbiter (ESA and Roscosmos), making open access to a vast collection of Mars surface images easy.

Numerous digital elevation models (DEMs) of Mars have been developed, each offering varying trade-offs in resolution and spatial coverage. The most widely used global dataset is the Mars Orbiter Laser Altimeter (MOLA) DEM \cite{mola}, produced from altimetry data acquired by the Mars Global Surveyor. MOLA provides nearly global coverage at a resolution of approximately 463 m/px. Global Mars's moon Phobos DEM generated by the Mars Express mission also exists, offering 100 m/px resolution.

To increase spatial detail, the High Resolution Stereo Camera onboard Mars Express has been used to create blended HRSC/MOLA DEMs with a typical resolution of 200 meters per pixel \cite{blended_mola}.

For further studies, requiring even higher resolutions, there exist DEM datasets derived from the CTX and HiRISE instruments' imagery aboard the Mars Reconnaissance Orbiter. Although the CTX imagery achieves nearly global Mars coverage, DEM generation with stereo-pipelines requires two images of the same location, significantly reducing the actual DEM coverage of Mars for that instrument. Moreover, DEMs generated from stereo pairs, may contain missing data or artefacts in places with low stereo texture or poor lighting conditions. The CTX and HiRISE-derived DEMs reach resolutions up to 18 m/px and 1 m/px respectively. Data repositories with CTX DEMs with HiRISE DEMs nested within are available \cite{day2023mars}.

Lastly, the Exo Mars's mission Trace Gas Orbiter is capable of generating DEMs from single passes with the CaSSIS instrument with stereo imaging capabilities. DEMs generated with the CaSSIS instrument can reach resolution up to 4.6 m/px, covering 0.1 \% of the Martian surface as of 2022.

While these datasets have enabled substantial progress in planetary science, they present several limitations for machine learning applications. Existing datasets are fragmented, often contain voids or noise that require substantial preprocessing and don't offer a clear way of dividing the data into parts without data leakage. As a result, researchers often need to implement custom pipelines to clean and repair elevation data before it can be used for training.

\subsection{CTX instrument}

The Mars Reconnaissance Orbiter (MRO), launched by NASA in 2005, focuses on exploring Mars' history of water, identifying potential landing sites for future missions, and acting as a critical relay for surface missions. Equipped with a suite of scientific instruments, MRO has provided invaluable data on Mars' climate, geology, and the presence of minerals that formed through interactions with liquid water. On board of MRO is the Context Camera (CTX), which captures high-resolution images of the Martian surface at ~6 meters per pixel, covering swaths about 30 km wide and up to 160 km long, depending on orbit and conditions. The CTX's primary purpose is to provide contextual imaging for more detailed observations from other instruments like HiRISE, helping to understand large-scale geological features and to map potential regions of interest across the planet.

\subsection{Monocular Depth Estimation}

Monocular depth estimation (MDE) is the task of inferring the relative or metric depth information about a scene from a single image. Current state-of-the-art methods for monocular depth estimation involve transformer-based architectures trained on synthetic and real images in a supervised and self-supervised manner, or leverage diffusion models that use an iterative noise-based process for data generation \cite{depthv1, depthv2, marigold}. One major limitation or bias present in these existing solutions is the fact that they have been primarily trained on "daily-life" images that include only a ground-based perspective and a distant background present within the frame of the picture, serving as a robust reference point \cite{depthv1}. DEM generation by monocular depth estimation has been explored by using generative adversarial networks (GANs) \cite{mars_remote_sensing_gans}. The authors design a network that simultaneously performs DEM generation, as well as super resolution of the input and the resulting DEM \cite{mars_remote_sensing_gans}.

Previous studies have evaluated the performance of MDE foundation models, particularly DepthAnythingV2, on RGB satellite imagery for the task of canopy height estimation. These evaluations were conducted both in a zero-shot setting and after fine-tuning the models specifically for the task  \cite{depth-any-canopy}. The authors have found that DepthAnythingV2 offered satisfactory zero-shot performance and achieved state-of-the-art performance after the cross-domain adaptation process. This marks the importance of wide availability of specific domain datasets for foundation model adaptation processes.

\subsection{Monocular depth estimation datasets}

MDE is a fundamental computer vision task with its own unique challenges. There exists a plethora of machine learning-ready datasets, suited for building MDE models. Most of the well-known depth estimation or 2D to 3D datasets \cite{cityscapes, kitti, nyuv2, rellis3d, sun3d, wildplaces} contain scenes primarily from the ground perspective in urban areas \cite{cityscapes, kitti}, frequently used in autonomous driving and navigation use cases, indoor settings \cite{nyuv2, sun3d} and sometimes natural environments \cite{wildplaces, rellis3d}. The common element between all of them is that they exclusively contain scenes with a ground-based perspective, which is understandable as mostlyxl industry applications make use of it. LiDAR instruments \cite{wildplaces, rellis3d}, time of flight instruments, such as Microsoft Kinect \cite{nyuv2, sun3d}, or stereo vision  \cite{cityscapes} are the most common ways of generating the ground truth in these datasets. This causes most of the currently available depth estimation datasets to be biased towards the ground-based perspective. The \datasetname dataset provides data for a similar task, but purely from an aerial perspective.

\section{Methods}
In this section, we detail the source of data used throughout the dataset generation process, the issues identified within, its origin and the pre-processing used to generate it. We describe in detail the structure of the generated dataset as well as the processing pipeline used to make it suitable for machine-learning use cases.
\subsection{Data source}
\begin{figure}
    \centering
    \begin{subfigure}[b]{0.38\linewidth}
        \includegraphics[width=\linewidth]{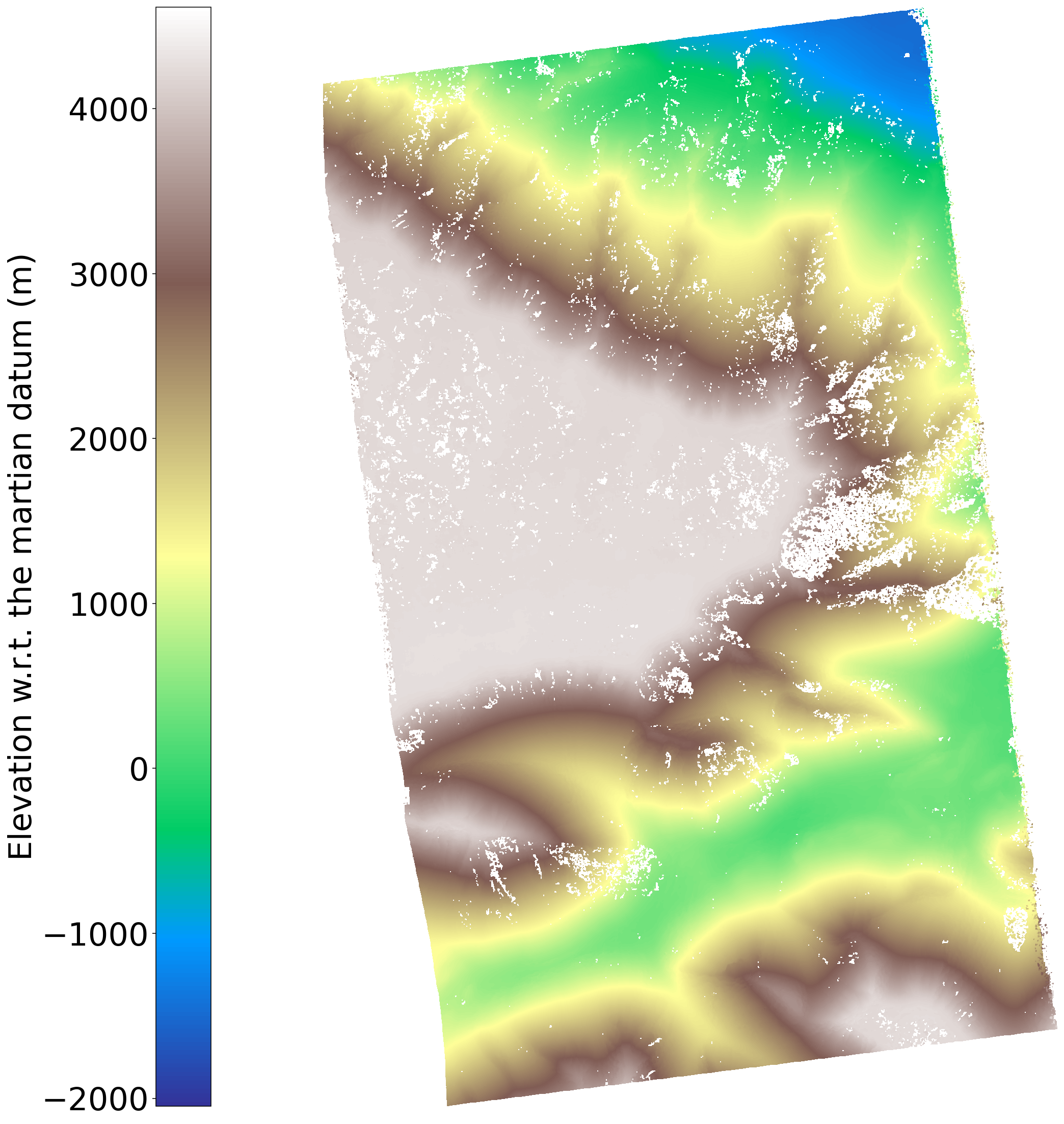}
        \caption{DEM}
        \label{fig:labeled_dem}
     \end{subfigure}
     \hfill
     \begin{subfigure}[b]{0.27\linewidth}
        \includegraphics[width=\linewidth]{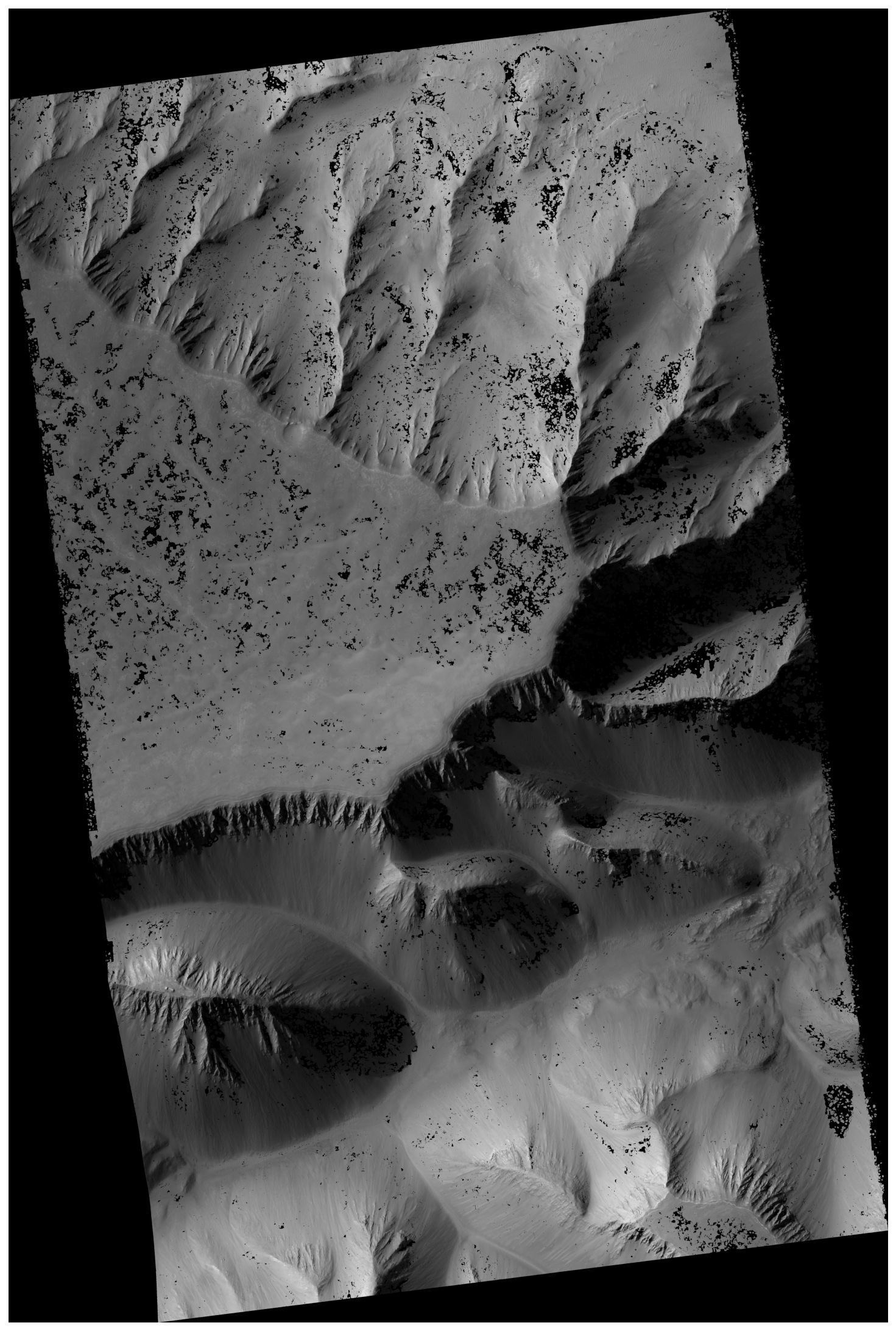}
        \caption{Orthoimage}
        \label{fig:labeled_optical}
     \end{subfigure}
     \hfill
     \begin{subfigure}[b]{0.27\linewidth}
        \includegraphics[width=\linewidth]{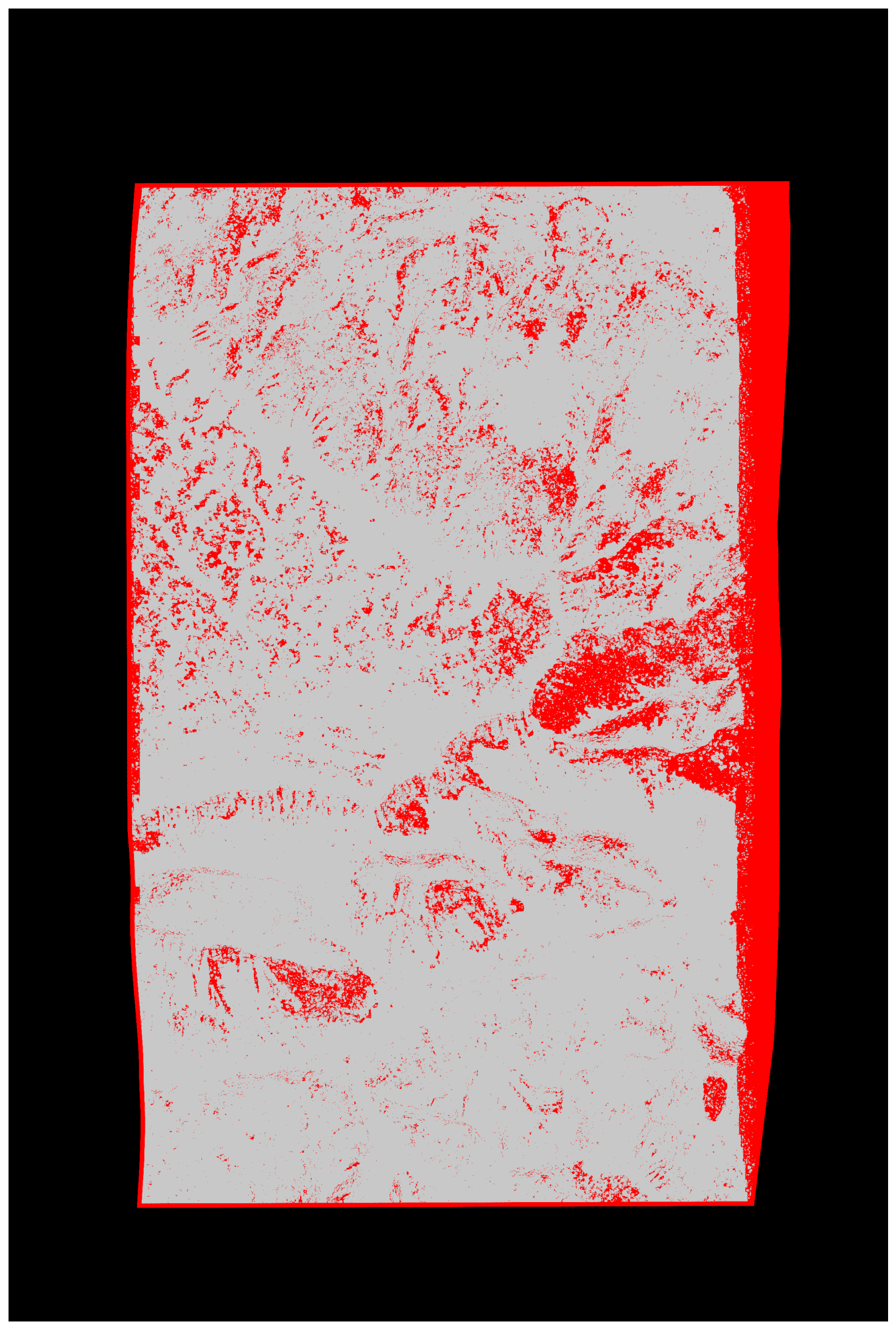}
        \caption{Goodpixelmap}
        \label{fig:labeled_mask}
     \end{subfigure}
     \caption{Every part of a single \citeauthor{day2023mars} sample available in the repository.\\  \small \citeauthor{day2023mars} sample name:  \texttt{b02\_010423\_1720\_xi\_08s084w\_b18\_016726\_1719\_xn\_08s084w}\normalsize}
     \label{fig:repository_data_types}
\end{figure}

\begin{figure}
    \centering
    \includegraphics[width=0.7\linewidth]{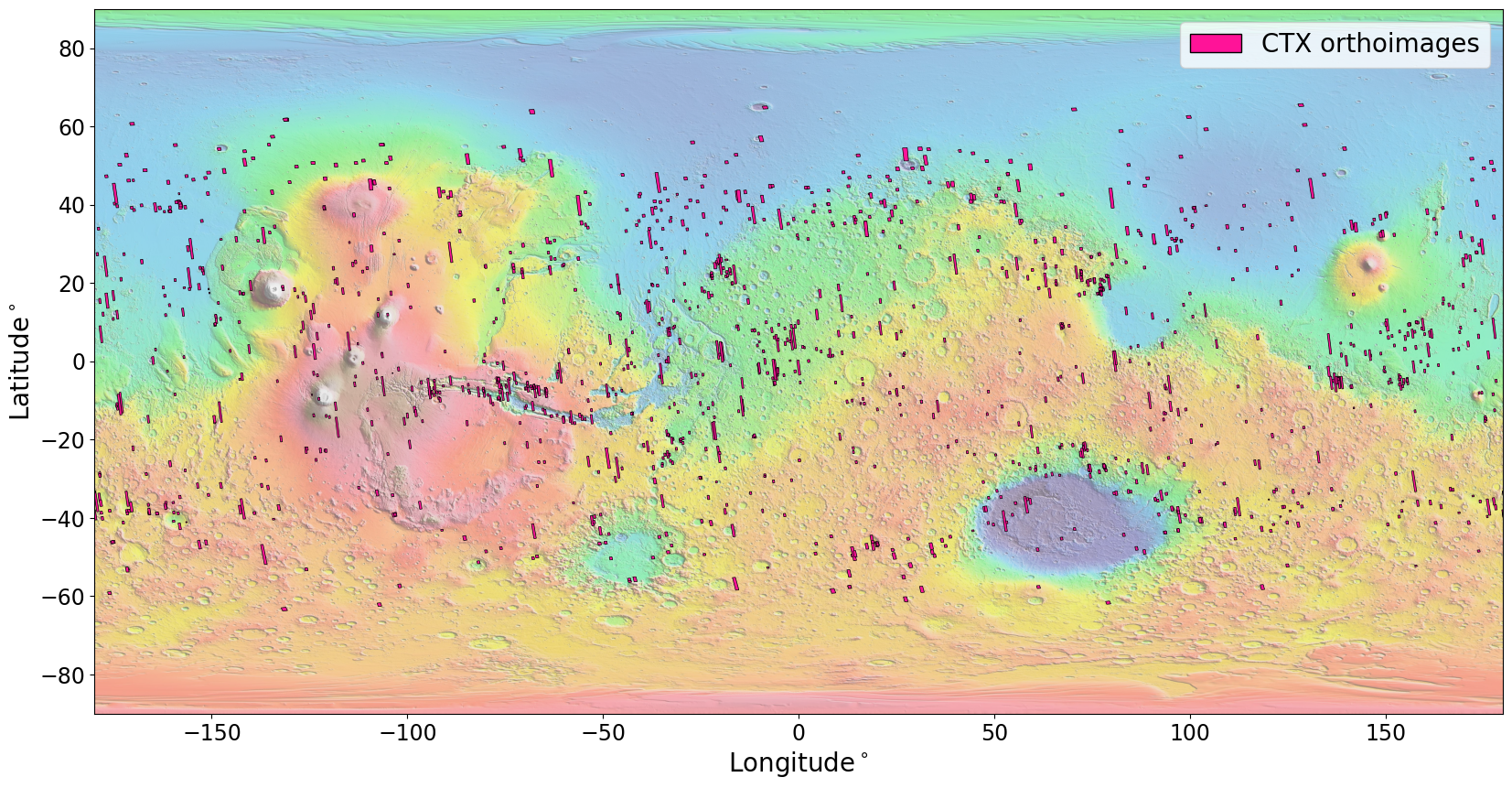}
    \caption{Localisation of all CTX-derived orthoimages and DEMs in the repository on the surface of Mars.}
    \label{fig:all_ctx_samples}
\end{figure}

Creating the \datasetname dataset, we utilised the repository by \citeauthor{day2023mars}, consisting of 1,354 orthoimage and DEM pairs generated from samples taken with the CTX instrument onboard the Mars Reconnaissance Orbiter and processed using the Ames Stereo Pipeline (ASP) \cite{AmesPipeline}. CTX imagery is a logical choice for the creation a global digital elevation model of Mars using monocular depth estimation as it has a global coverage of Mars, at high resolution and the images are relatively consistent in lighting and resolution \cite{dickson}, which are crucial for monocular depth estimation techniques that rely on uniform image characteristics to infer depth accurately. The creation of a DEM using the ASP requires two images of the same terrain captured from slightly different perspectives, and at its heart lies the process of pixel-to-pixel correlation in both of the images. This process, for certain areas, may not end successfully, making the derivation of elevation data from such a point impossible. This may cause some generated elevation maps to be incomplete, and realistically, most generated DEMs have at least some data points missing. To address this issue, the authors supply each pair with an additional mask file marking all points for which the correlation has been unsuccessful (files with the \texttt{goodpixelmap} suffix). All of the data types are shown in Figure \ref{fig:repository_data_types}. The repository covers numerous regions of scientific interest on Mars, including the Jezero Crater, the Gale Crater, and Oxia Planum, along with many other areas distributed on the Martian surface. The exact distribution of the samples gathered in the entire repository has been showcased in Figure \ref{fig:all_ctx_samples}.

\subsection{Data quality}
While the \citeauthor{day2023mars} repository is a valuable asset, the data it contains present several challenges, limiting its usability in machine learning model training. We identified the following shortcomings in the repository contents: 
    \paragraph{\textbf{Misaligned masks}} Both the orthoimages and the DEMs contain missing values, due to the limitations of the pipeline they have been generated with. To address this, as mentioned previously, the authors provided a separate mask file called \texttt{goodpixelmap} for every sample, indicating for which values the correlation process has been successful. The mask at times has a different orientation than both the orthoimage and DEM file, requiring a proper alignment before superimposition. As the mask is meant to mark individual pixels for which the correlation has been unsuccessful, the alignment should also be of pixel-wise accuracy, greatly reducing its usability.
    
\begin{figure}
    \centering
    \begin{subfigure}[b]{0.35\linewidth}
        \includegraphics[width=\linewidth]{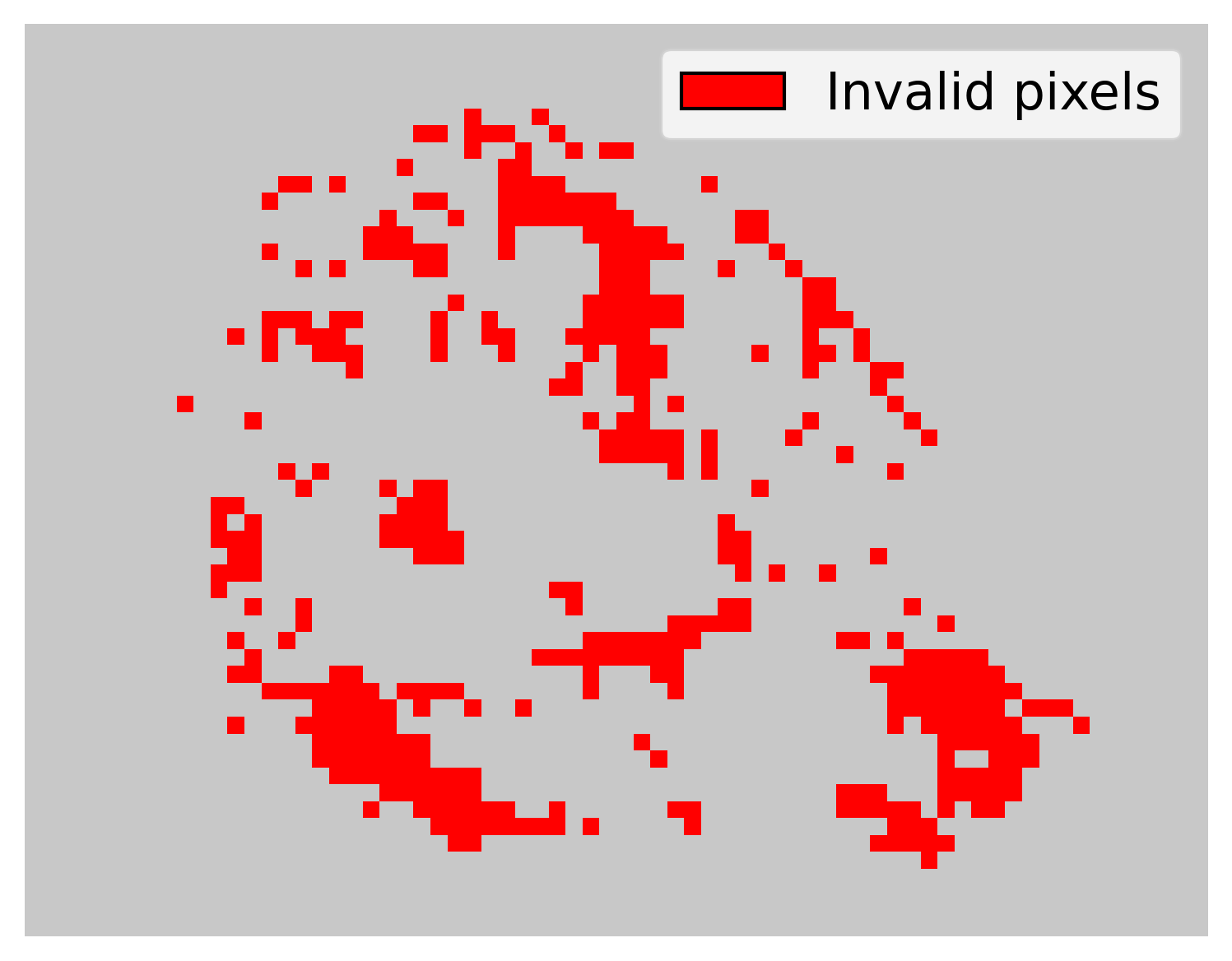}
        \caption{Fragment of the original \texttt{goodpixelmap}}
        \label{fig:misalignment_mask}
     \end{subfigure}
     \hfill
     \begin{subfigure}[b]{0.23\linewidth}
        \includegraphics[width=\linewidth]{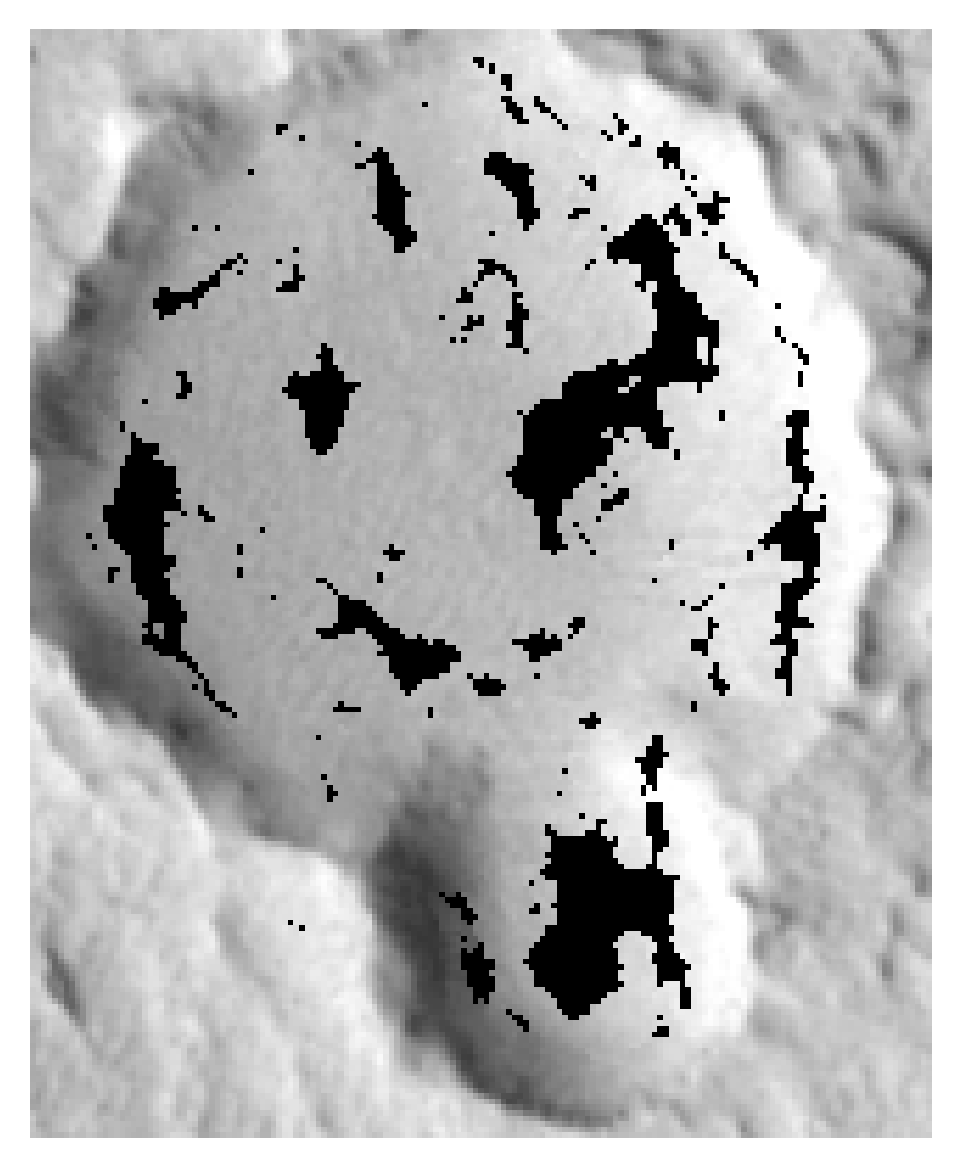}
        \caption{Fragment of the original optical image}
        \label{fig:misaligned_image}
     \end{subfigure}
     \hfill
     \begin{subfigure}[b]{0.39\linewidth}
        \includegraphics[width=\linewidth]{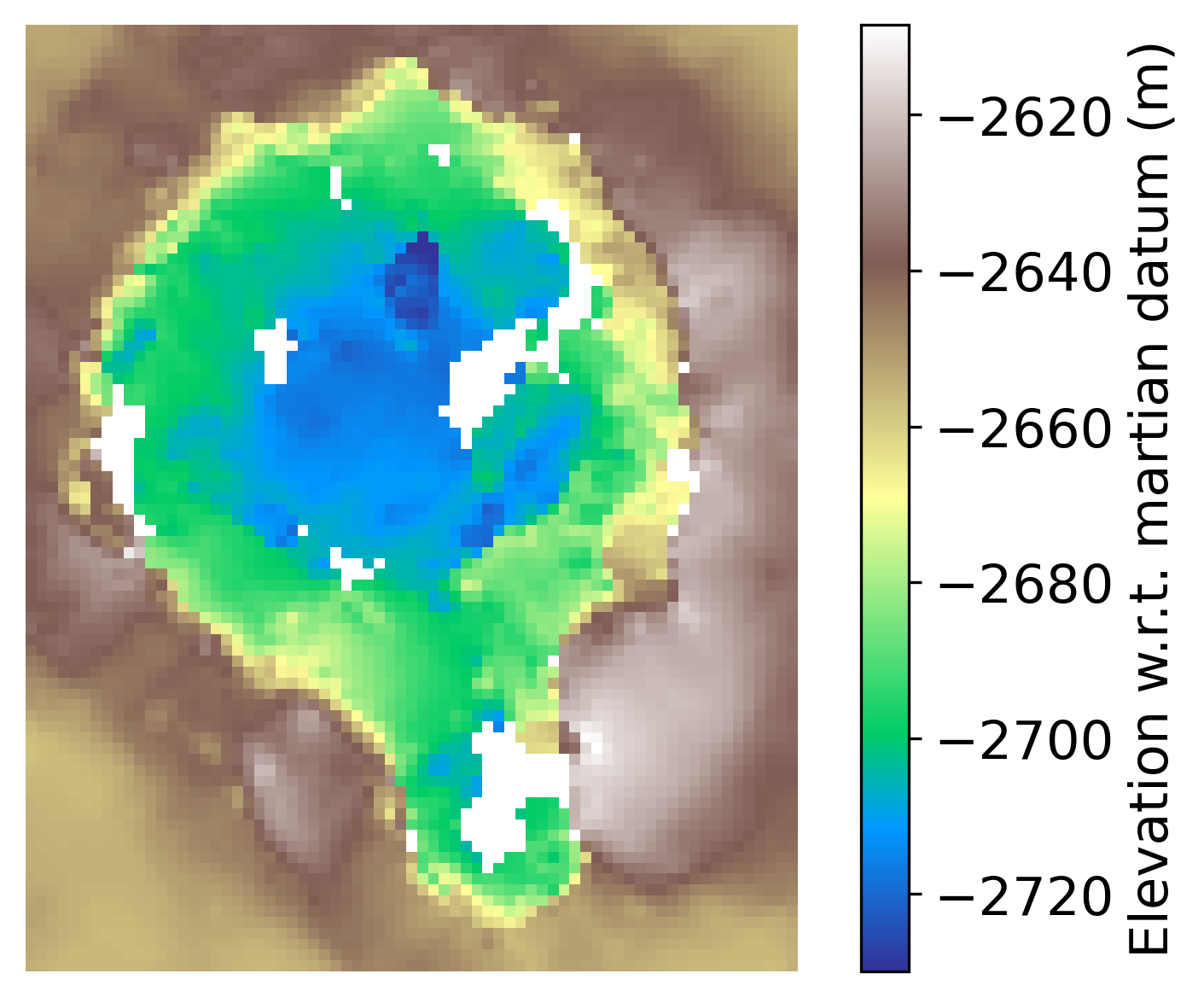}
        \caption{Fragment of the original digital elevation model}
        \label{fig:misaligned_dem}
     \end{subfigure}
     \caption{The same terrain fragment as it appears in all three data types in the original form. Discrepancies between the resolutions and orientation can easily be seen. The data types require co-registration and re-scaling to a common resolution in order to superimpose them. Because of the different native resolutions of each data type, it's impossible to exactly match all of the features on each data type; all of them represent a different level of detail that is irreversibly lost in lower resolutions. This mismatch between the data types makes precise superimposition impossible.\\ \small \citeauthor{day2023mars} sample name: \texttt{b07\_012511\_1819\_xn\_01n211w\_b07\_012234\_1820\_xn\_02n211w}\normalsize}
     \label{fig:misalignments}
\end{figure}

    \paragraph{\textbf{Different resolutions}} The orthoimage, mask and the DEM typically all have different resolutions, requiring either upscaling or downscaling of the rest of the data types. Both transformations are irreversible, either removing or interpolating new information. This inevitably causes compatibility and alignment issues even when all of the data types have been verticalised and brought to the same resolution. This has been visualised in Figure \ref{fig:misalignments}. Due to this quality issue, we decided to completely omit using the provided masks in the dataset generation process.

\begin{figure}[!htbp]
    \centering
     \begin{subfigure}[b]{0.3\linewidth}
     \centering
    \includegraphics[width=\linewidth]{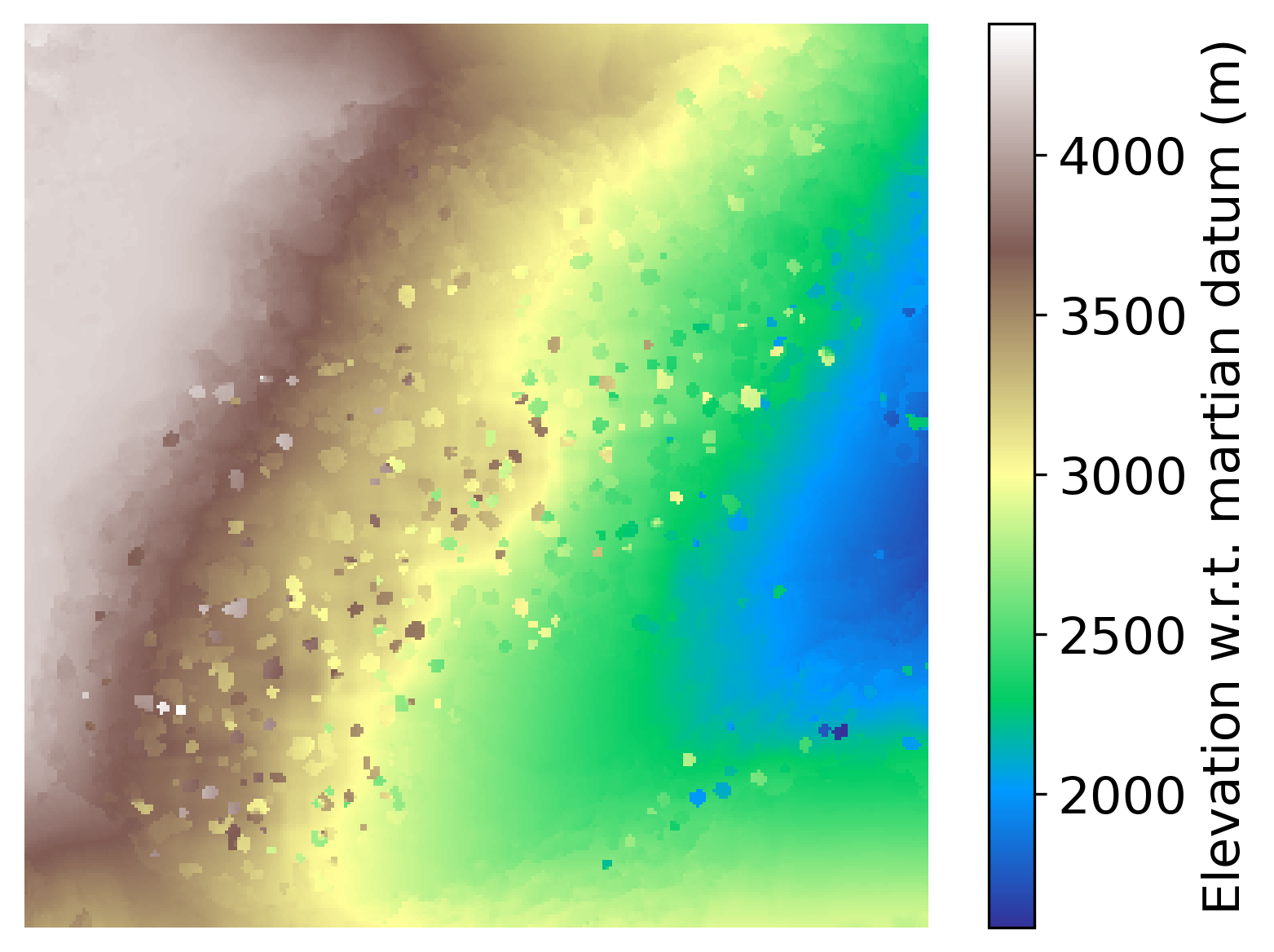}
    \caption{Elevation value artefacts visible as small "islands" scattered within a large area where elevation values were filled in.}
    \label{fig:illustration_islands}
    \end{subfigure}
    \hspace{0.05\linewidth}
    \begin{subfigure}[b]{0.3\linewidth}
    \centering
    \includegraphics[width=\linewidth]{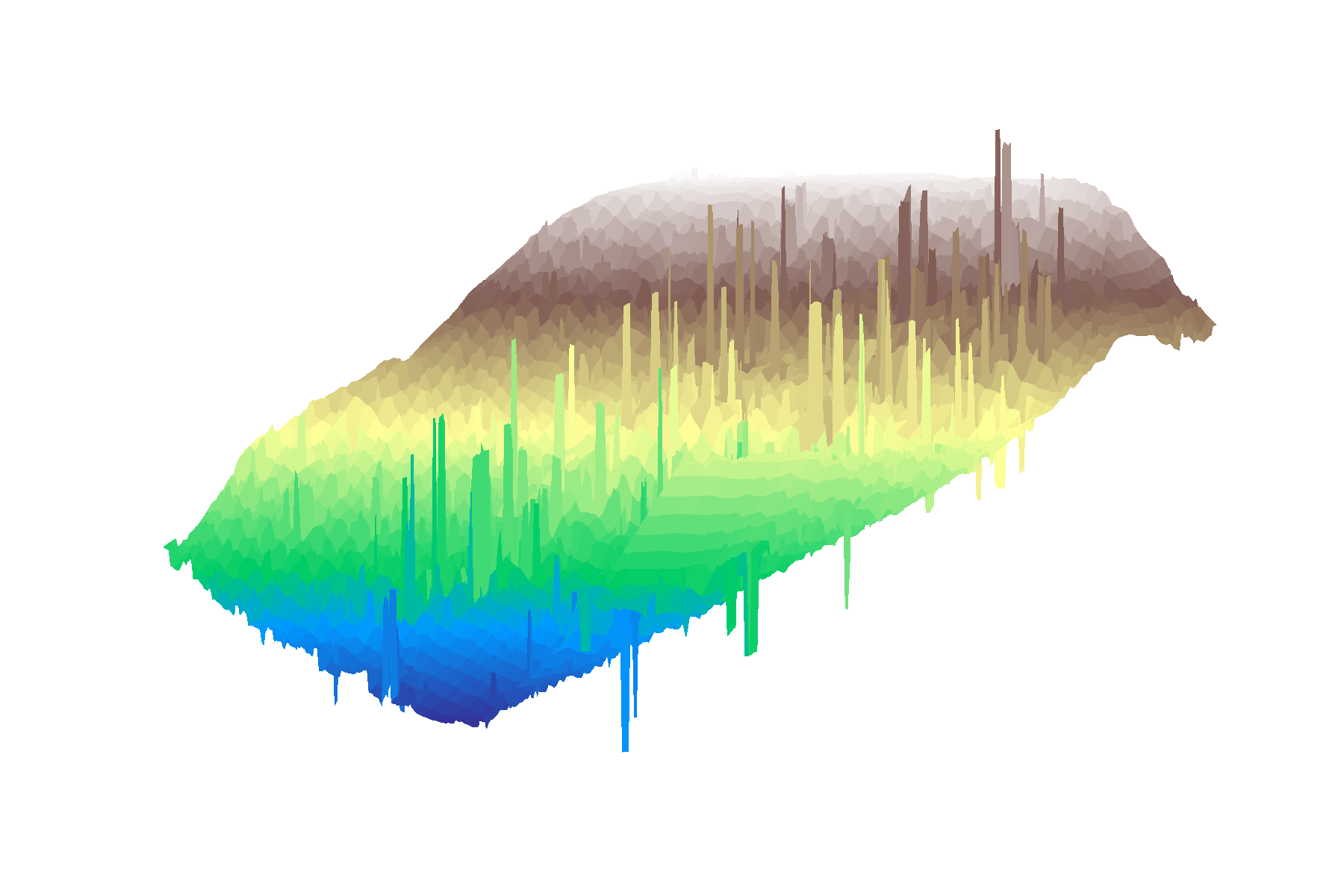}
    \caption{Elevation value artefacts visible as steep spikes or wells on a 3D view of the terrain.}
    \label{fig:illustration_islands_3d}
    \end{subfigure}
    \caption{Elevation artefacts on elevation maps in the repository}
    \label{fig:elevation_artifacts}
\end{figure}

    \paragraph{\textbf{Elevation artefacts}} Some of the regions of the generated DEMs are undefined, due to the limitations of the ASP. We frequently observed large areas with undefined elevation values with small "islands" of valid elevation data scattered within (Figure \ref{fig:illustration_islands}). Those pocketed islands typically exhibit extreme elevation values, detached from the nearest large valid elevation areas, appearing as high spikes or deep wells on the elevation map (Figure \ref{fig:illustration_islands_3d}). This is most likely caused by these points missing local elevation references during generation.

\paragraph{\textbf{Distorted samples}} Some of the samples in the repository are heavily distorted, appearing as stretched or missing many of elevation values. Examples are shown in Figure \ref{fig:distorted_samples} in the Appendix \ref{appen}. Due to the required severe readjustment, these samples have been omitted in the dataset generation process.\\

\begin{figure}[!htb]
    \centering
    \includegraphics[width=0.4\linewidth]{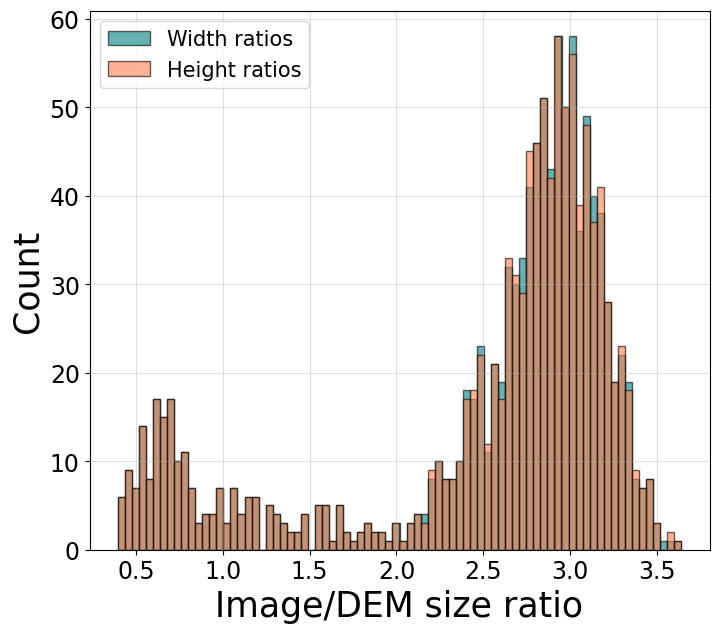}
    \caption{Histograms of the height and width ratios between the optical images and DEMs in the \citeauthor{day2023mars} repository. Image resolutions are roughly three times the size of the DEMs in most cases.}
    \label{fig:height_width_ratios}
\end{figure}

Another fact worth noting is that authors of the \citeauthor{day2023mars} repository claim the CTX-based DEMs' resolution is equal to 18m/px. Since the resolution of optical CTX images is around 6m/px, we compared the heights and widths of the optical images with the same dimensions in their DEM counterparts. The results can be seen in Figure \ref{fig:height_width_ratios}. It is evident that most samples fall in the vicinity of the optical image being thrice the resolution of the DEM, which would agree with the authors' claim, although the spread is quite substantial. Non-negligible number of samples do not follow the same rule; therefore, it cannot be said that the dataset we provide as a result of this work supplies DEMs consistent with the 18m/px resolution.\\

For some shortcomings, we have developed processing pipelines to mitigate their impact; for some, we simply rejected the data samples. This, in turn, brought the total number of used samples from 1,353 to 1,122 samples.

\subsection{\datasetname}
\textbf{Mars CTX Terrain Elevation Dataset} (\datasetname) is a dataset of paired orthoimage and DEM patches, tailored for machine learning purposes, created from an extensive repository \cite{day2023mars} of orthoimages and DEMs derived from images taken by the CTX instrument onboard the Mars Reconnaissance Orbiter NASA mission. We developed data processing and curation pipelines and conducted a preliminary level of exploratory analysis. The dataset was created to fill the gap between depth estimation and remote sensing datasets, as we see the DEM generation as a derivative of the depth estimation task. Each sample in the dataset consists of four different parts. These include:

\begin{itemize}
    \item \textbf{Orthoimage patch (uint8)} - a 518x518 pixels patch of the original orthoimage, intended to serve as the input data. The images are monochromatic, but are still supplied as RGB images, with all channels containing the same values. We chose this particular patch size as it's the input size of common MDE foundation models like DepthAnythingV2 \cite{depthv2}.
    \item \textbf{DEM patch (float32)} - the corresponding elevation map, indicating the elevation of each point in the optical orthoimage. The elevation patches don't contain any invalid elevation values. It has the same size as the orthoimage patch.
    \item \textbf{Invalid elevation mask (bool)} - a binary mask indicating which pixels in the elevation map were invalid in the \citeauthor{day2023mars} data sample before processing.
    \item \textbf{Elevation outlier mask (bool)} - a binary mask indicating pixels that have been considered as elevation anomalies, excluding invalid values, during processing.
\end{itemize}

\begin{figure}[h]
    \centering
    \includegraphics[width=0.9\linewidth]{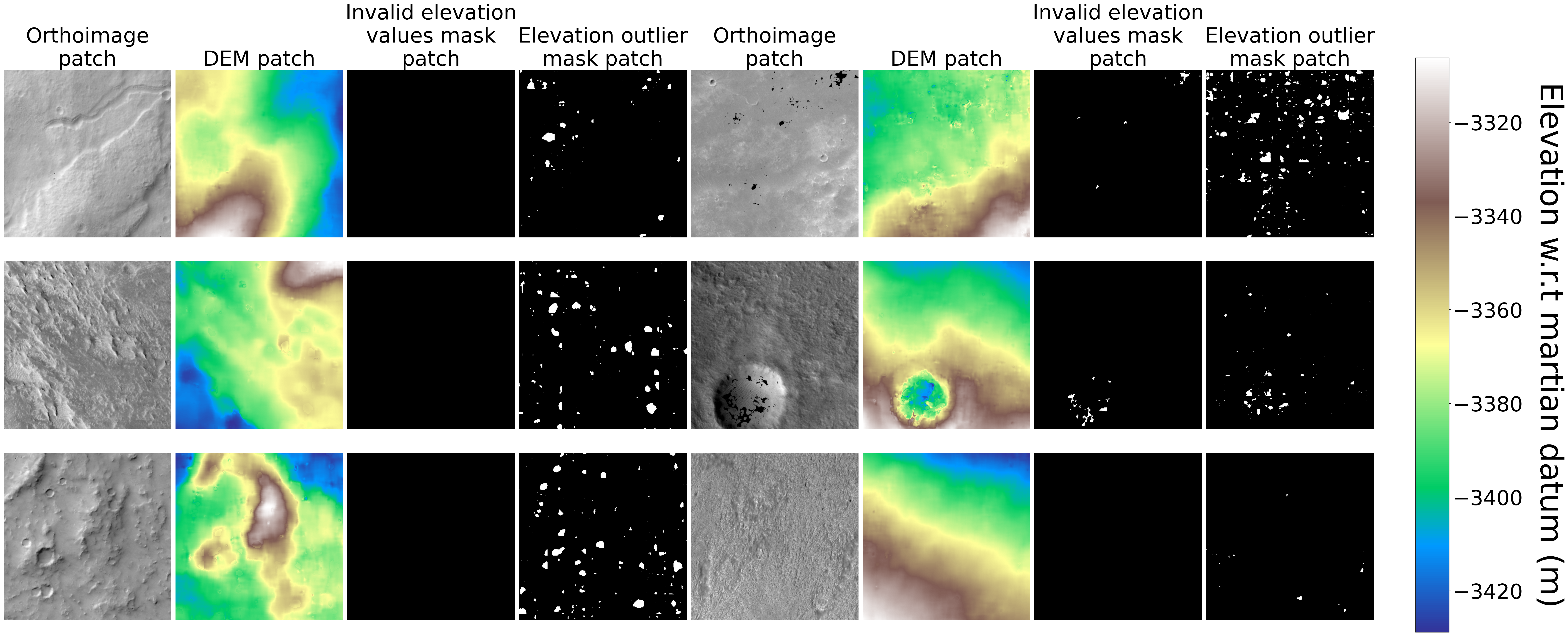}
    \caption{Showcase of 6 different samples from the dataset, each sample is represented by the optical orthoimage, DEM patch and the invalid values and elevation outlier masks.}
    \label{fig:dataset_samples}
\end{figure}

Examples of all of the data types for several samples from the dataset have been shown in Figure \ref{fig:dataset_samples}.

\subsection{Data processing}

To create the \datasetname dataset, we developed an extensive processing pipeline to address the mentioned quality issues and transform the data to a form ready to be used for machine learning tasks. The pipeline consists of both image and elevation map processing steps, as well as input and output data sample selection processes. Figure \ref{fig:processing_diagram} in the Appendix \ref{appen} illustrates the processing pipeline implemented for processing the dataset. Below, we describe each step in detail.
\begin{enumerate}
    \item \textbf{Day et al. sample selection} - at the beginning of the pipeline, a candidate sample for processing may be rejected because of two reasons. Some samples in the repository are heavily distorted (Figure \ref{fig:distorted_samples}), which can be detected by looking at their aspect ratio. Valid samples coming from the CTX instrument tend to be spanned along the south-north direction. A complete histogram of aspect ratios for the repository is shown in Figure \ref{fig:aspect_ratios}. We apply a hard aspect ratio threshold of $\frac{W}{H}=1$, rejecting samples exceeding that threshold, as below that threshold is the highest concentration of non-distorted samples. Another way for sample rejection before processing may be due to pixels exceeding set values for both images and elevation maps.

\begin{figure}
    \centering
    \includegraphics[width=0.7\linewidth]{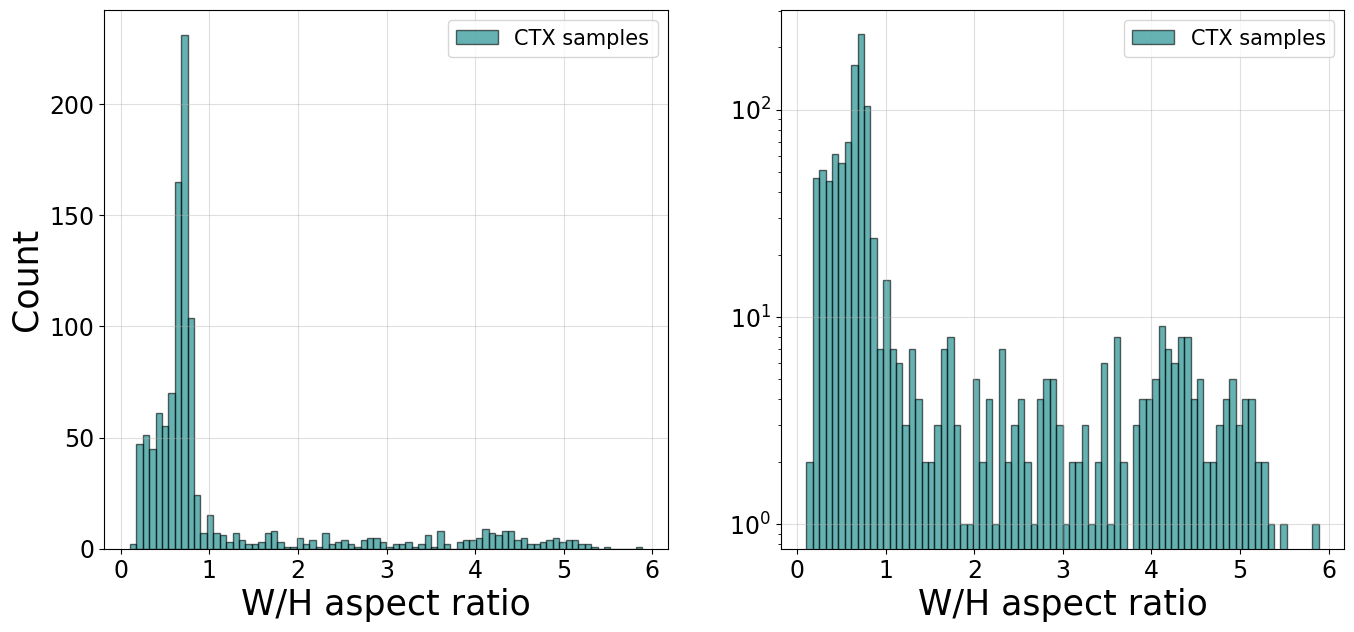}
    \caption{Histograms of $\frac{W}{H}$ aspect ratios for the CTX samples in the repository. The right figure shows the histogram with a logarithmic scale.}
    \label{fig:aspect_ratios}
\end{figure}

    \item \textbf{Preliminary map filling} - the first step of the processing involves filling in all of the invalid elevation values before performing verticalisation of the elevation map. Rotating the elevation map may require interpolating new elevation values. Leaving the invalid values in place before performing rotation will result in corrupting the elevation values in the nearest vicinity of the invalid value due to the needed interpolation.

    \item \textbf{Verticalisation} - before further processing, the samples are verticalised to save on processing time and avoid pixels outside of the terrain boundary influencing the result. The detailed description of how the verticalisation was implemented can be found in the Appendix \ref{appen}.
    
    \item \textbf{Missing value binary mask generation} - invalid elevation values in the repository are indicated in the elevation maps with the value $z=-32767.0$. We generate a binary mask, marking the location of pixels equal to that value. This is needed to correctly fill in the gaps in elevation data.

    \item \textbf{Filling in missing elevation values} - after the binary mask is created, the missing elevation values need to be populated. We calculate each missing elevation value as the average value of its selected valid neighbourhood of a chosen size.

    \item \textbf{Elevation outlier binary mask generation} - as described previously, elevation maps from the repository suffer from the presence of elevation artefacts, highly prominent in small valid elevation value islands within large gaps of missing elevation data (Figure \ref{fig:elevation_artifacts}). To address this issue, we employ a sliding window filter of a set size that compares each value within that window to the window's median $\tilde{h}$. Additionally, we weigh every pixel within that window with a 2D Gaussian function centred in that window with a parametrised spread $\Sigma$. We calculate the standard score $Z(x,y)$ w.r.t. the window's median $\tilde{h}$ and standard deviation $\sigma_h$ for every pixel and consider pixels above a parametrised threshold $T$, elevation outliers.
    \begin{equation}
    m_o(x,y) = \left\{\begin{matrix}
     1,& Z(x,y)> T  \\
     0,& Z(x,y)\leq T \\
    \end{matrix}\right.,
    \qquad
    Z(x,y) = w_g(x,y)\frac{h(x,y) - \tilde{h}}{\sigma_h},
    \qquad
    w_g(x,y)\sim \mathcal{N}(\boldsymbol{0}, \boldsymbol{\Sigma})
    \end{equation}\\
    where $x$ and $y$ are coordinates within the sliding window. This effectively creates a binary mask with marked elevation artefacts $m_o$.

    \item \textbf{Join initial missing elevation mask and elevation outlier mask} - the two binary masks are combined using the \textbf{OR} logical operator, achieving a new mask of all elevation locations that need to be filled in again. This is necessary as some of the values filled in initially would be computed using values coming from the elevation artefacts. With this mask, created the pipeline simply jumps back to \textbf{Step 3} and the process is repeated a set number of times, using different neighbourhood sizes and deviation thresholds to refine the completed elevation map. The missing value and outlier binary masks are kept and saved together with the image and elevation map, as they may be considered during validation.

    \item \textbf{Patching} - After processing the masks, images and elevation map are divided into equally sized chunks we call patches.

    \item \textbf{Patch selection} - after the sample has been processed by the pipeline, each patch is passed through another selection process. This aims at eliminating patches that contain too many black pixels in the optical images, elevation patches that are too flat or elevation patches with too many imputed elevation values:
    \begin{enumerate}
        \item \textbf{black pixel content} - we reject patches containing more than 10\% of purely black pixels on the optical patch. Images mostly or entirely black make it impossible to infer the correct elevation value, as no features can be analysed on black images.

        \item \textbf{imputed elevation values content} - if the contribution of imputed elevation values for a given patch is above 15 \%, it is also rejected. We set a requirement on the patches that are included in the dataset to contain a certain amount of non-altered elevation values, ensuring most data in the dataset is not augmented.

        \item \textbf{flat elevation patches} - we aim to avoid including overly flat terrain into the dataset, therefore we set a requirement of a minimal standard deviation of each elevation patch to $\sigma_h = 10\text{m}$. Patches that don't meet this requirement are rejected.
        
    \end{enumerate}    
\end{enumerate}

\begin{figure}[h]
    \centering
    \includegraphics[width=0.6\linewidth]{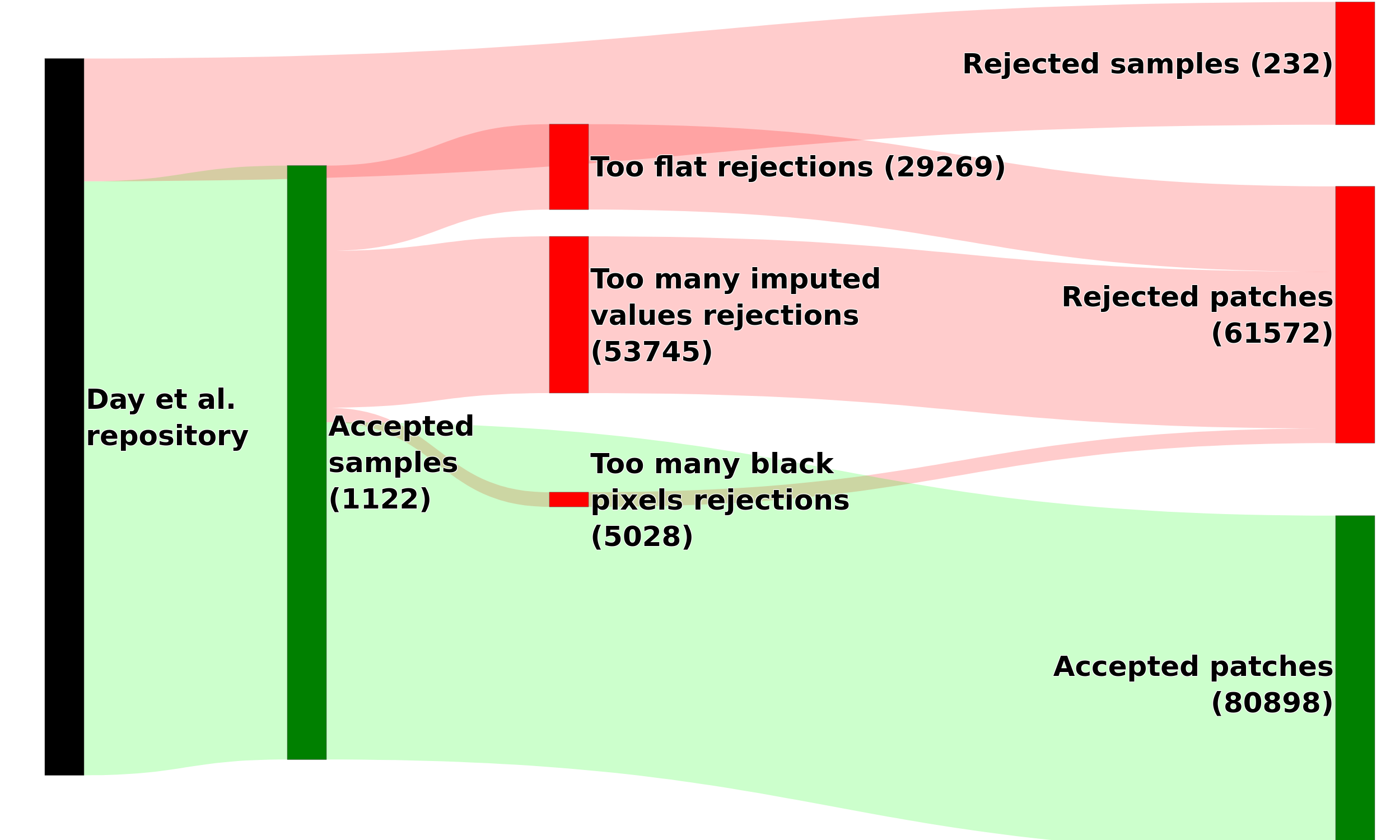}
    \caption{Sankey diagram of how we processed the \citeauthor{day2023mars} repository. The patch rejections sum up to a greater number than the number of rejected patches, due to some patches meeting multiple rejection conditions. The size of bars is proportional to the amount of data at each step.}
    \label{fig:sankey_processing}
\end{figure}

The detailed breakdown of how much data from the \citeauthor{day2023mars} repository we were able to use for the creation of the dataset has been visualised in Figure \ref{fig:sankey_processing}.

The entire processing pipeline is configurable with a single \texttt{yaml} file, containing all of the processing parameters. We derived the final parameter values by manually adjusting them and inspecting the processing output in real time. While choosing the parameters, we looked for the correct identification of elevation outliers while minimising the false positives. This process of manual adjustment has been conducted for 54 patches coming from four \citeauthor{day2023mars} samples with a high number of artefacts. For the final set of parameters, the most frequently occurring values have been chosen. The detailed set of parameters used for \datasetname dataset generation has been included in the Appendix \ref{appen}.

\subsection{Dataset split generation}

\begin{figure}
    \centering
    \includegraphics[width=0.7\linewidth]{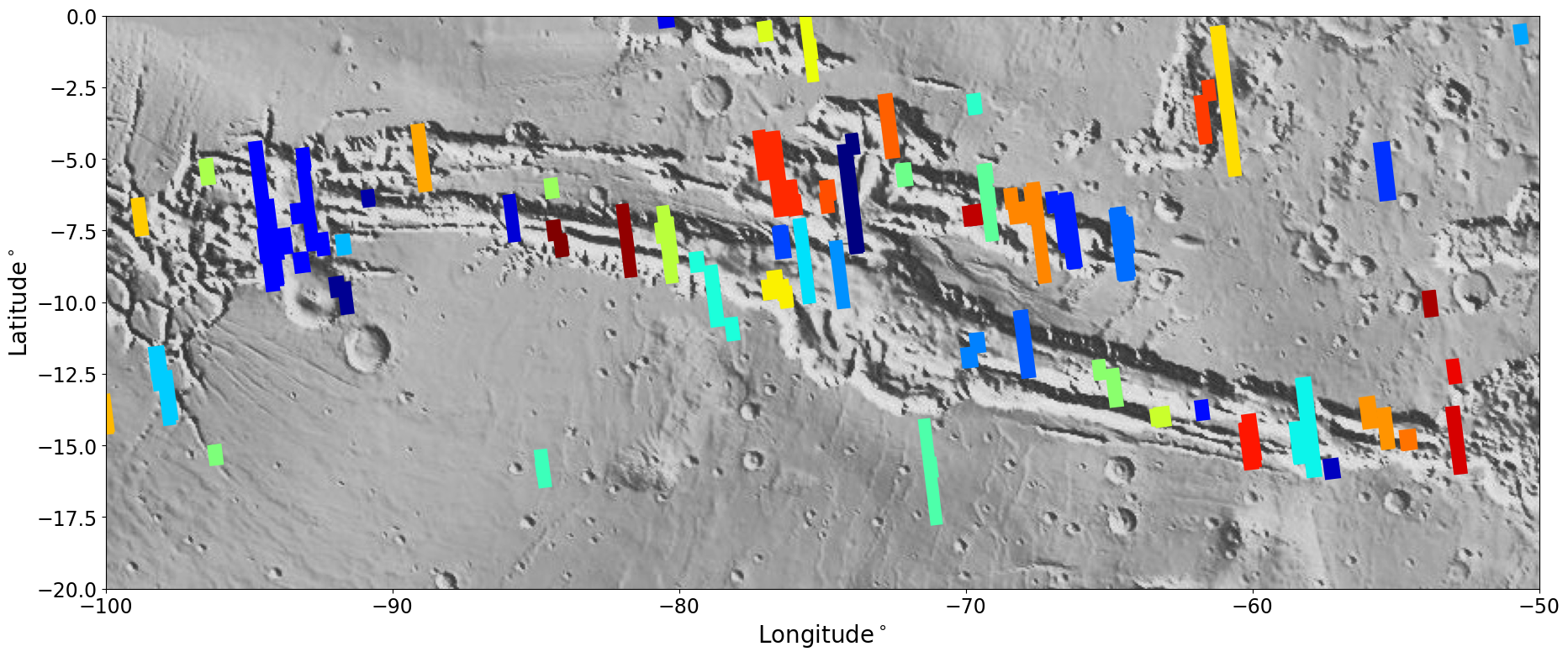}
    \caption{Visualisation of sample clusters around the Valles Marineris area, as it's among the most sample-dense regions. Each separate cluster is represented by a unique colour.}
    \label{fig:clusters_visualization}
\end{figure}

\begin{figure}
    \centering
    \includegraphics[width=0.7\linewidth]{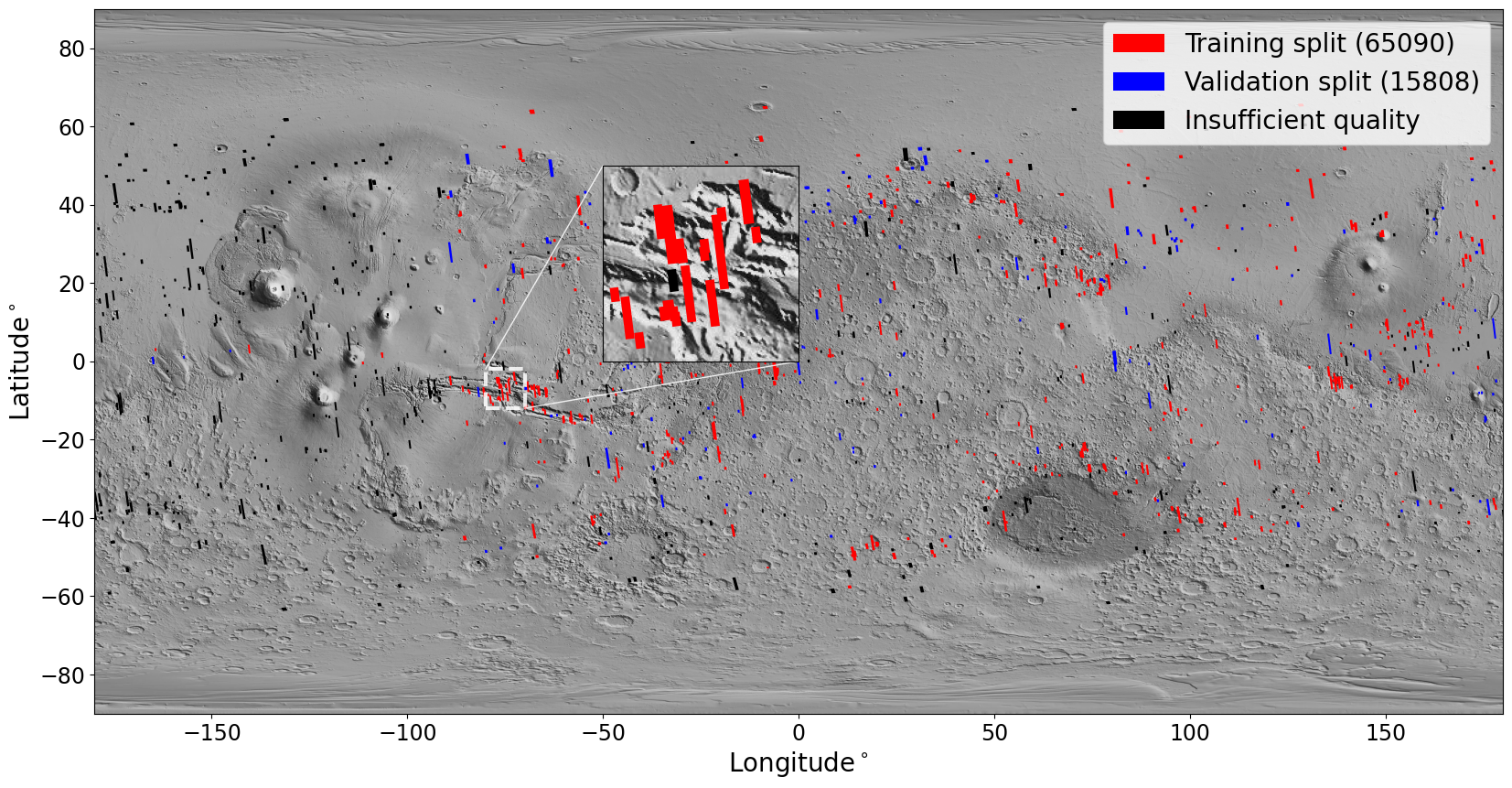}
    \caption{Clusters divided into training and validation splits. During the initial sample selection or the following patch selection, samples are discarded, due to that some samples end up not yielding any patches. These samples have been marked with black. Many of samples between $[-180^\circ, -100^\circ]$ latitude have been discarded.}
    \label{fig:dataset_split}
\end{figure}

We supply the \datasetname dataset in a form typical for machine learning, which is a division into a training and validation split. Proper division of the dataset into training and validation splits must ensure that no data that has been used in model training ends up in the validation split. In this case, that means no location imaged on a patch in the training set can appear in the validation split. To ensure there is no data leakage between the training and validation splits, we make use of the bounding boxes for every swath used in the process of generating the stereo pairs. All of the bounding boxes for the left images are shown in Figure \ref{fig:all_ctx_samples}. Since the features in the resulting orthoimage can come from either the left or right images, we create unions of the left and right image bounding boxes for each orthoimage. We then search recursively for all intersecting unions and assign them to sample clusters, with all overlapping samples belonging to the same cluster. This guarantees that no images belonging to one cluster intersect with images from a different cluster. We then divide the generated patches into splits cluster-wise, meaning all patches from one cluster always end up in the same split. The visualisation of derived clusters is shown in Figure \ref{fig:clusters_visualization}, and a clear division of clusters into dataset splits is shown in Figure \ref{fig:dataset_split}.

\begin{table}[h]
    \caption{Number of data samples in each dataset split}
    \centering
    \begin{tabular}{ll}
    \toprule
        Training split & Validation split \\
        \midrule
         65090 & 15808 \\
    \bottomrule
    \label{tab:dataset_splits_numbers}
    \end{tabular}
    
\end{table}

\section{Results}
\subsection{Statistical Overview}
We conducted a thorough investigation into different aspects of the generated dataset, like elevation values, slopes and sample spatial distribution, to aid its future users by providing an introductory statistical overview, as well as to ensure the validation split is representative of the training data. In this section, we provide various figures and quantitative measures, showing these aspects of the \datasetname dataset. 

\begin{figure}[h]
    \centering
    \includegraphics[width=0.7\linewidth]{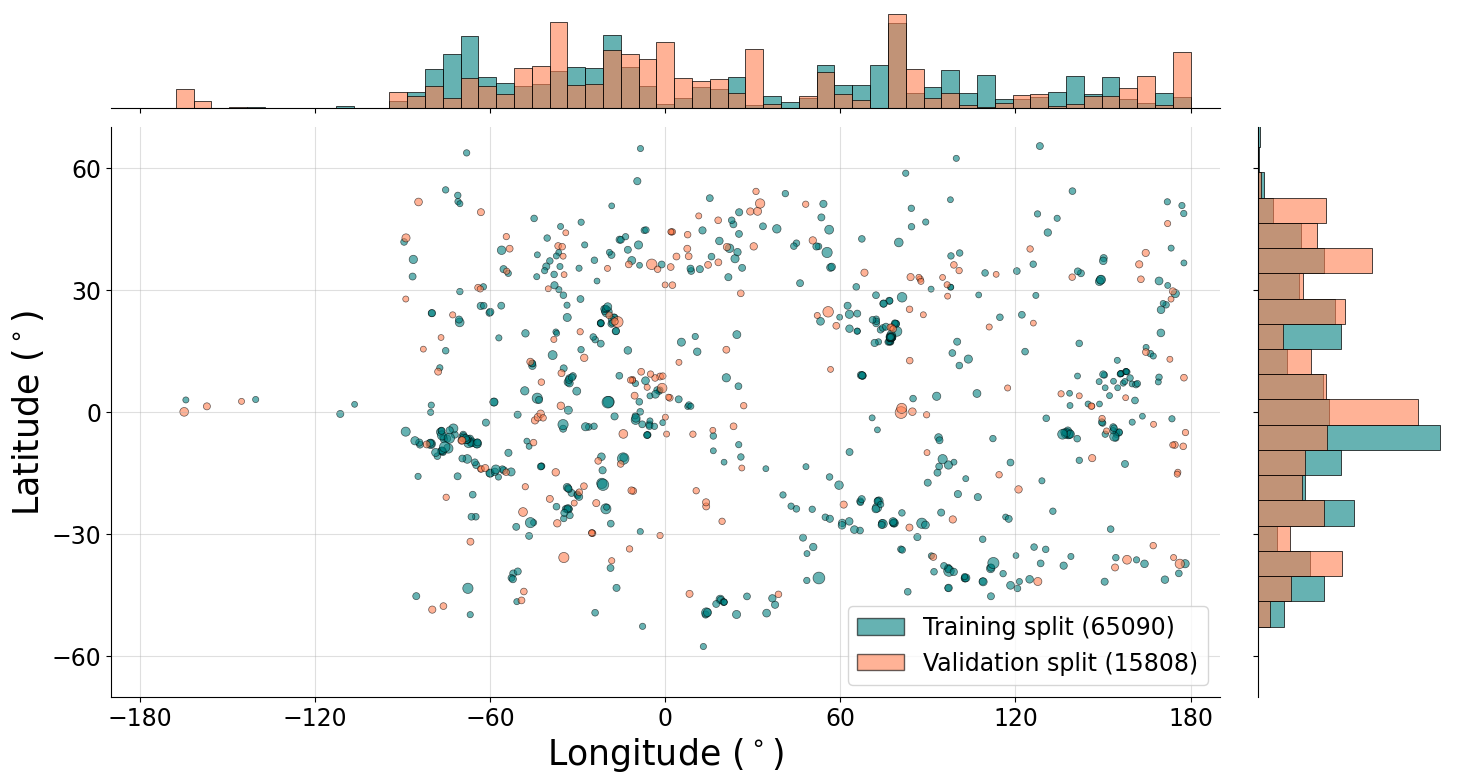}
    \caption{Spatial distribution of samples between the splits, the size of the data points is proportional to the number of patches coming from a particular sample. The lack of samples from the $[-180^\circ, -120^\circ]$ is visible.}
    \label{fig:spatial_distribtion_splits_2d}
\end{figure}

\begin{figure}[h]
    \centering
    \includegraphics[width=0.7\linewidth]{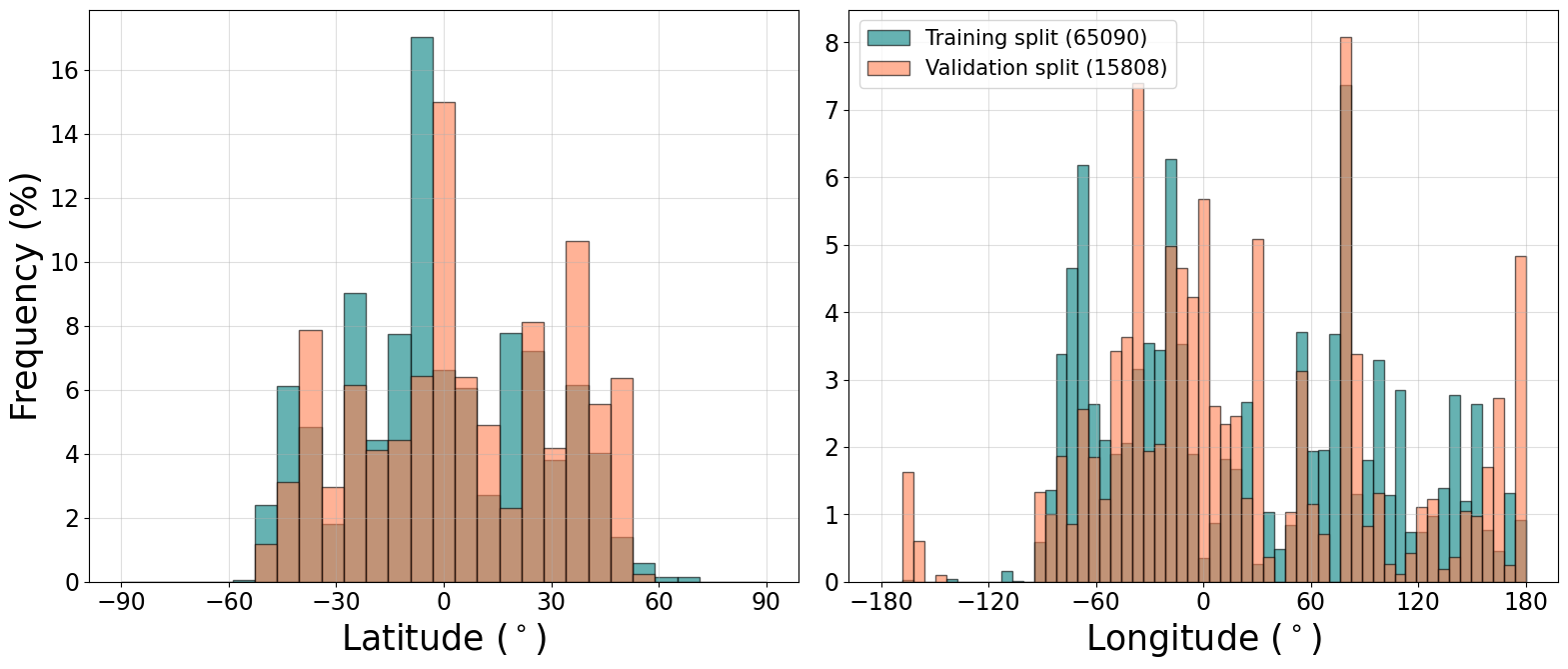}
    \caption{Latitudinal and longitudinal distributions of patch frequency across both dataset splits.}
    \label{fig:lat_long_dists}
\end{figure}

\paragraph{Spatial distribution}
The repository from \citeauthor{day2023mars} contains samples that overall provide a good coverage of the Martian surface, beyond the polar regions. In Figure \ref{fig:spatial_distribtion_splits_2d}, the spatial distribution of samples has been shown, showing a high degree of similarity. Distributions of latitude and longitude for data samples from the original repository used in the generation of the dataset are shown in Figure \ref{fig:lat_long_dists}. Most of the used samples come from a rectangular area spanning between $[-120^\circ, 180^\circ]$ longitude and $[-60^\circ, 60^\circ]$ latitude values. As mentioned previously, the majority of samples from areas with longitudes in $[-180^\circ, -120^\circ]$ range, had insufficient quality to be used in the dataset generation process.

\begin{figure}
    \centering
    \includegraphics[width=0.7\linewidth]{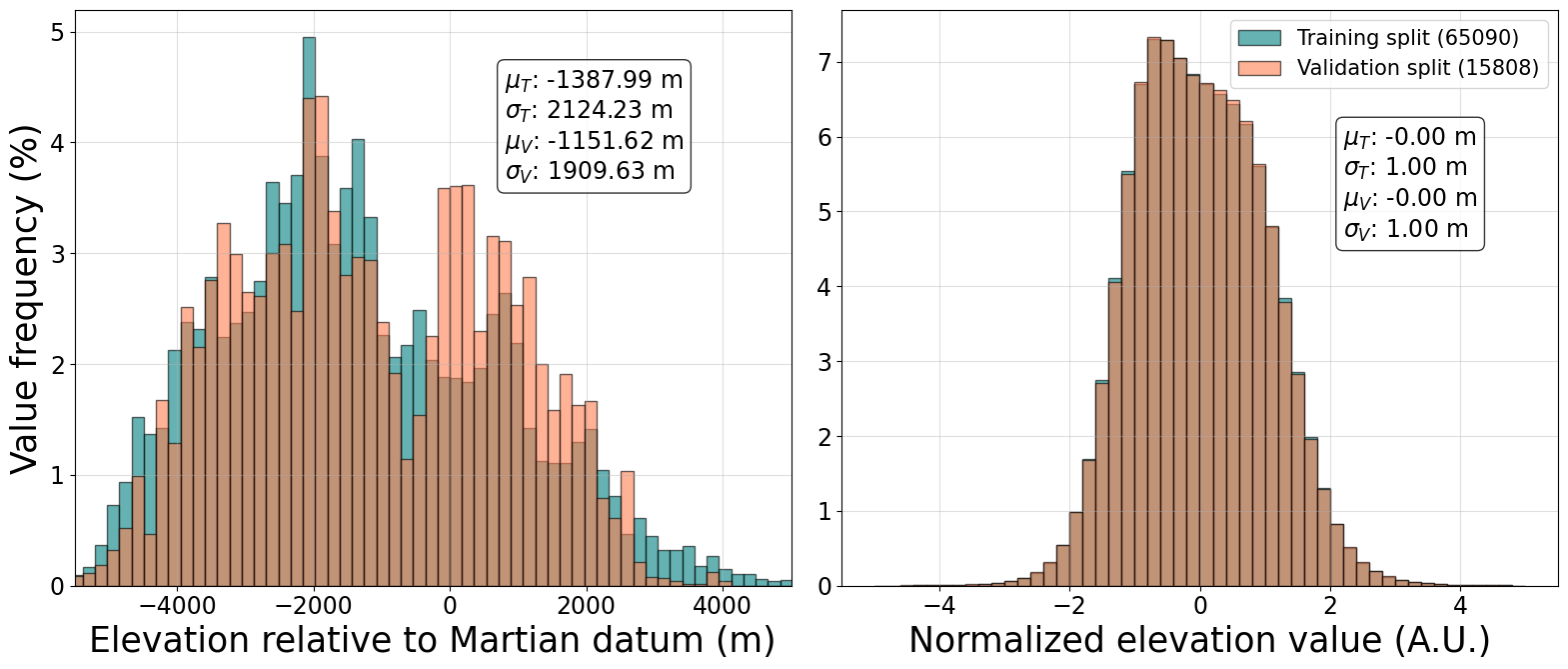}
    \caption{Distributions of elevation values across both datasets splits in the original and standardised space. The histogram has been truncated to elevation values up to 5,000 m, as this is where the majority of the values are contained. There are some elevation values as high as 10,000 m. The histograms were created using 10,000 randomly sampled elevation values from every patch ($\sim$3.7\% of pixels from every patch). As expected, the standardised elevation distributions are almost identical, both exhibiting a slight skew of the distribution towards negative relative depth values. The metric elevation value distributions are largely similar with slight differences due to some regions being either largely a part of only one split (Valles Marineris) or being poor in data samples, resulting in the only available samples inevitably being a part of only one of the splits (Hellas Planitia).}
    \label{fig:elevation_distributions}
\end{figure}

\paragraph{Elevation distributions}
We provide the patches with the elevation values in an unchanged form, spanning a range from -5,000 to 10,000 meters with respect to the Martian datum. The distribution of elevation values across all patches in the dataset has been shown in Figure \ref{fig:elevation_distributions} in both original and normalised space. For monocular relative depth estimation, we convert each patch into standardised space, masking invalid pixels to not be taken into account during that process.

\begin{figure}
    \centering
    \includegraphics[width=0.4\linewidth]{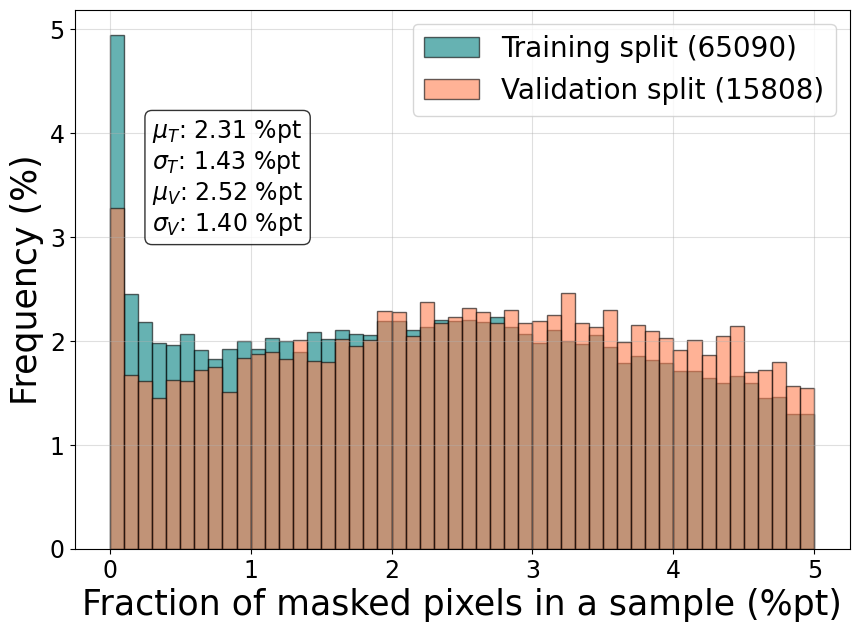}
    \caption{Distribution of masked pixels frequency in both dataset splits.}
    \label{fig:masks_distributions}
\end{figure}

\paragraph{Masked pixels distribution}
In the dataset, we included two types of binary masks that indicate modified or filled-in values in the elevation maps. These masks can be used to ignore errors that may originate in places where the data has been altered, allowing for training models only on non-altered data. The distribution of masked pixel frequency in patches has been shown in Figure \ref{fig:masks_distributions}.

\paragraph{Slopes distribution}
Lastly, we provide a distribution of mean slopes for both splits of the dataset. The mean slope aims to visualise how steep or flat the elevation maps are across the dataset.

\begin{figure}
    \centering
    \includegraphics[width=0.7\linewidth]{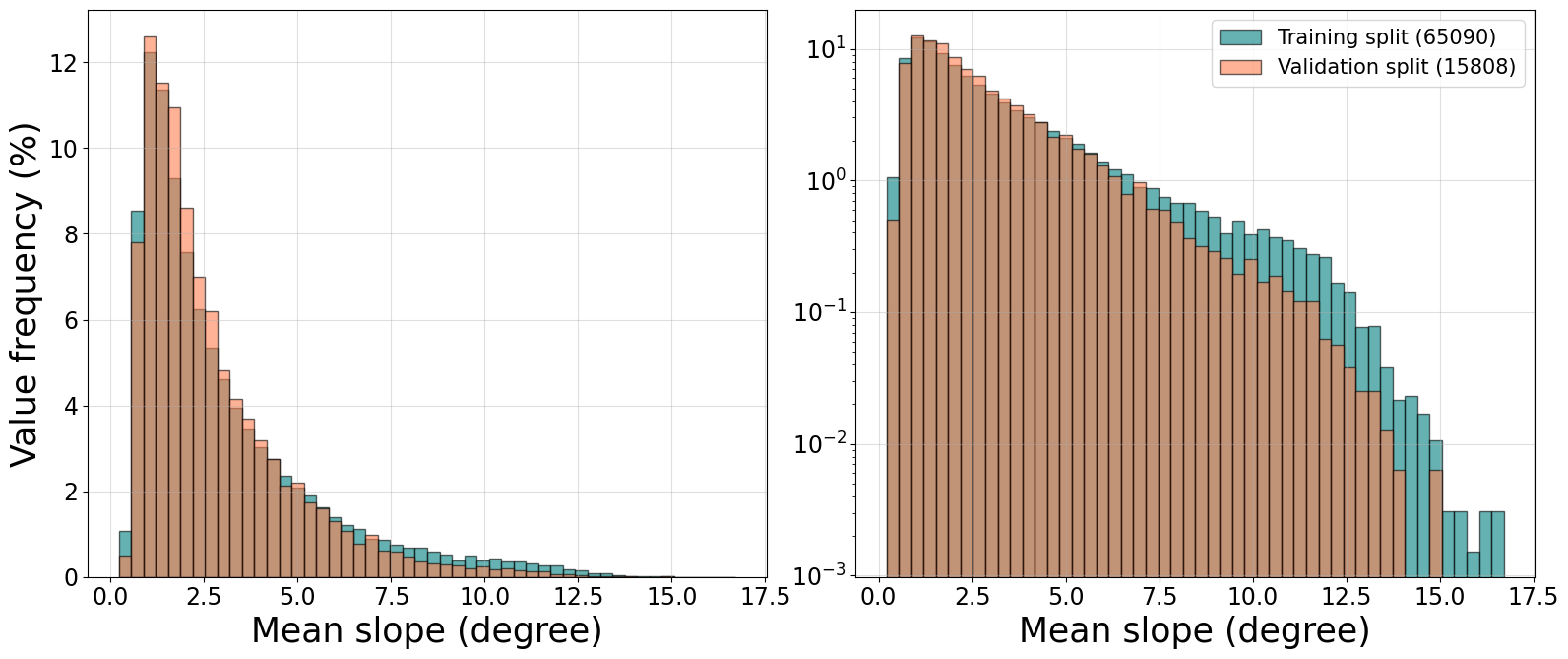}
    \caption{Distribution of the mean elevation patch slope across the entire dataset in linear and logarithmic scale.}
    \label{fig:slopes_distribution}
\end{figure}

\subsection{Baseline}
We evaluate a simple U-Net architecture on the task of DEM generation using the provided dataset. This aims at establishing baseline performance metrics, as well as determining if the dataset is suitable for achieving a reliable elevation prediction on unseen data.

\subsubsection{Experiment results}
We train the baseline model on the DEM prediction task formulated as a relative depth estimation task. We standardise each DEM patch and treat this normalised representation as relative depth. We then calculate the mean squared error (MSE) between the prediction and ground-truth patch to estimate the reconstruction loss during training.

\paragraph{Model architecture}
We employ a simple U-Net architecture with three compression and expansion stages. In the expansive path of the model, we use bilinear up-sampling layers instead of transposed convolutional layers. A crucial modification for this task was the complete replacement of batch normalisation layers with instance normalisation layers. When batch normalisation was used during training, the model tended to converge at a trivial solution of generating a flat plane in the vicinity of the dataset's mean elevation. The detailed structure of the model can be found in the Appendix \ref{appen}.

\begin{figure}[!htb]
    \centering
    \includegraphics[width=0.6\linewidth]{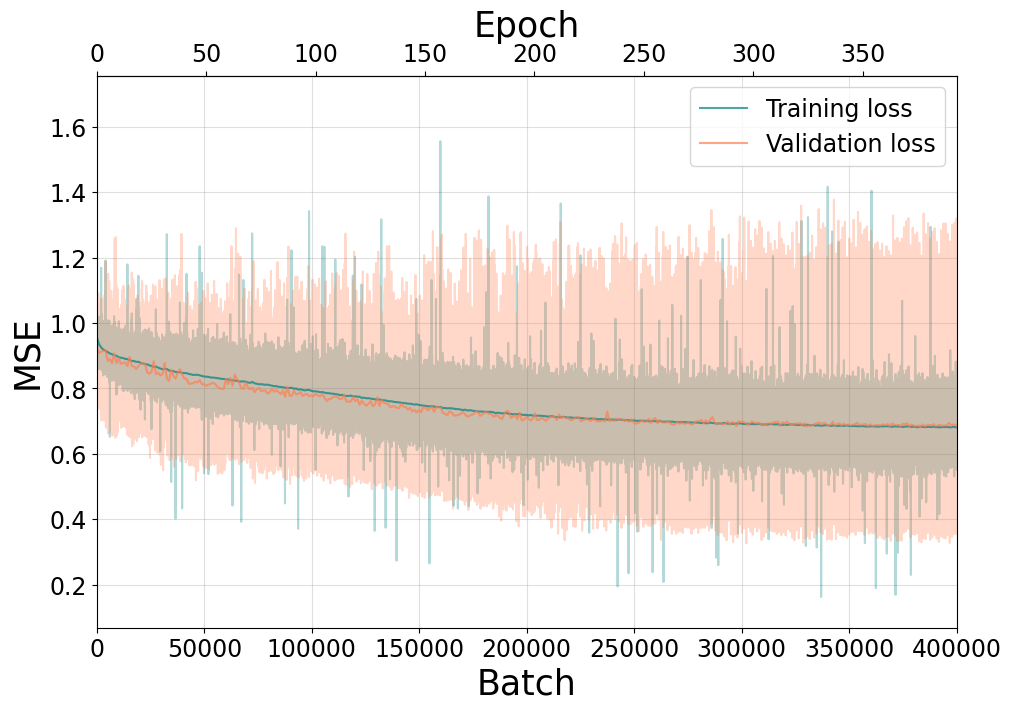}
    \caption{Training and validation loss curves during the training process up till 400,000 batches. For evaluation, we picked the checkpoint saved after the 300th epoch, whereupon visual inspection, the loss curves appear to intersect. The training continues with both losses eventually plateauing.}
    \label{fig:loss_curves}
\end{figure}

\paragraph{Model training}
We use the Adam optimiser during training, with MSE as the loss function. Due to working on one standardised dataset, we didn't find it necessary to rely on shift and scale invariant losses, which is commonly done in depth estimation tasks. The model is trained for 1,000 epochs with a batch size of 64. We use a base learning rate $\alpha=1\text{e-}4$ with a linear warm-up period of 20 epochs, starting at 10\% of the base learning rate, followed by an exponential decay with rate of $\gamma = 0.99$. During training, we employ a series of geometric and pixel-level augmentations to improve the model's robustness and prevent overfitting. We save the model checkpoint every 10 epochs, choosing for evaluation one that is close to the crossing point of the validation and training loss curves. We chose this approach due to the lack of a test split in the dataset, as we didn't want to evaluate a model that is optimised for performance on the validation split. With this approach, we picked a checkpoint saved after the 300th epoch. The corresponding loss curves are shown in Figure \ref{fig:loss_curves}.

\paragraph{Performance comparison}
We compare the performance of the baseline model with DepthAnythingV2 \cite{depthv2} monocular depth estimation foundation model on the task of DEM prediction. Both models generate their outputs in relative depth space $h_R$, with the DepthAnythingV2 model requiring an additional output normalisation step, as it was originally trained on multiple datasets with a scale and shift invariant loss
\begin{equation}
    h_R = \frac{h_{DAv2} - \mu_{h_{DAv2}}}{\sigma_{h_{DAv2}}}
\end{equation}\\
Our baseline model is trained to already generate predictions in that space. The predictions in standard score space are then rescaled to the original metric space of the ground truth data, using the mean $\mu_{GT}$ and standard deviation $\sigma_{GT}$ of the corresponding ground truth patch.
\begin{equation}
h_p = \sigma_{GT}\cdot h_R + \mu_{GT}
\end{equation}
\\
We then compare both models' performance, evaluating their prediction error using mean absolute error (MAE), root mean squared error (RMSE), relative error (RelErr) and relative absolute error (RelAbsErr). The relative errors are calculated against the range of the elevation values in the ground truth patch, instead of corresponding pixel values, as they may cross the zero point, causing the error to explode in that elevation region.
\begin{equation}
    \text{RelErr} = \frac{h_{GT} - h_p}{\text{max}(h_{GT}) - \text{min}(h_{GT})}
\end{equation}
\\
\begin{equation}
     \text{RelAbsErr} = \frac{\big|h_{GT} - h_p\big|}{\text{max}(h_{GT}) - \text{min}(h_{GT})}
\end{equation}
\\
The evaluation is performed using the validation split of the dataset. We compute all errors as an average metric for every patch, but to provide more insight into the evaluation, we also sample 100 locations from every patch to visualise how the errors are distributed. The average performance of both models is shown in Table \ref{tab:performance}.

\begin{table}[h]
    \caption{Mean evaluation metrics for both models.}
    \centering
    \begin{tabular}{lllllll}
    \toprule
        & RMSE & MAE  & RelErr & RelAbsErr & parameters \\
        \midrule
         U-Net baseline & 61.90 & 42.38 & -2.2e-4 & 0.13 & $\sim$500k\\
          \midrule
         DepthAnythingV2 & 123.34 & 79.97 & 2.85e-4 & 0.22 & $\sim$300M\\
    \bottomrule
    \end{tabular}
    \label{tab:performance}
\end{table}

\begin{figure}[!htb]
    \centering
    \includegraphics[width=0.6\linewidth]{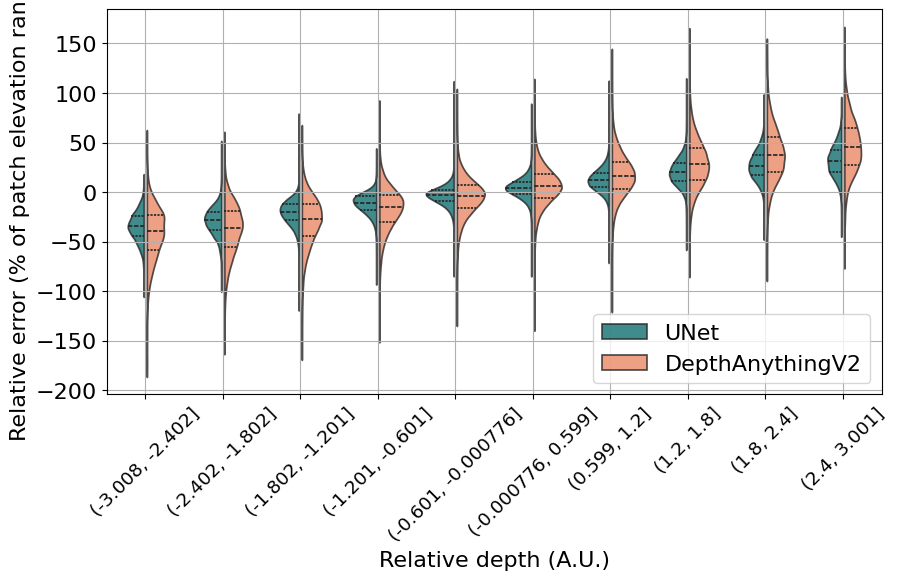}
    \caption{A comparison of relative error distributions across different ground-truth elevation ranges for both models. Errors represented here are for single pixels in a given elevation range. Both models are more accurate in generating predictions for a symmetric relative depth values range, apparent by the narrower distributions in the middle of the figure, centred around 0\% relative error.}
    \label{fig:error_violin}
\end{figure}

\begin{figure}[!htb]

    \centering
    \begin{subfigure}[b]{0.49\linewidth}
    \includegraphics[width=\linewidth]{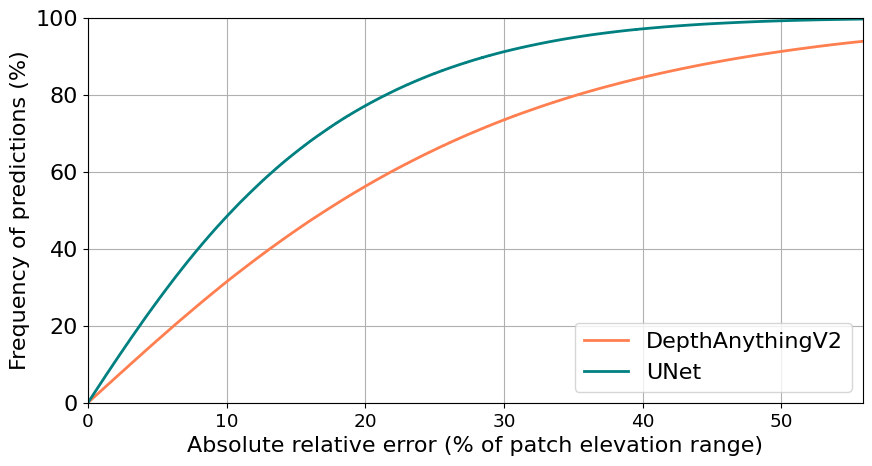}
    \caption{Cumulative distribution function (CDF) of absolute relative errors on the DEM prediction task for both models.}
    \label{fig:models_cdf}
    \end{subfigure}
    \hfill
    \begin{subfigure}[b]{0.49\linewidth}
    \includegraphics[width=\linewidth]{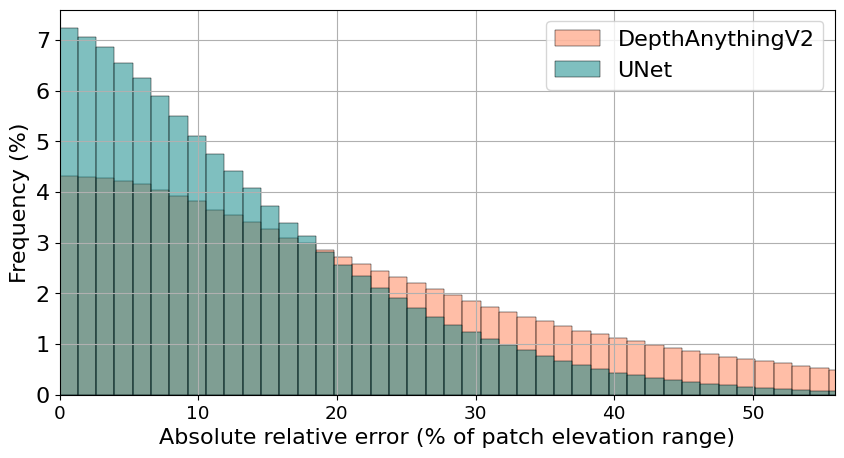}
    \caption{Histogram showing the distribution of absolute relative errors for both models across the validation split.}
    \label{fig:error_distributions}
    \end{subfigure}
    \label{fig:eval_comparisons}
    \caption{Comparison of absolute relative error distributions for both evaluated models, further showing the baseline model's predictions exhibiting smaller errors.}
\end{figure}

The small baseline model outperforms DepthAnythingV2 across all tested metrics, with the baseline offering roughly two times smaller evaluation error, while being approximately 600 times smaller in terms of number of parameters. This indicates that the massive parameter count, in conjunction with extensive pre-training on depth prediction tasks from the ground perspective, doesn't transfer well to the DEM generation task. This highlights that DEM generation, despite the apparent similarity, represents a significant domain shift for typical MDE models. DEMs generated by using the outputs predicted by the DepthAnythingV2 end up having a greater error variance. Additionally, we visualise the distribution of the errors w.r.t. to the elevation value, and compute the cumulative distribution function of the relative errors for both models, to further show the performance gap. These results are shown in Figure \ref{fig:error_distributions} and Figure \ref{fig:models_cdf}. As visible in Figure \ref{fig:error_violin}, both models exhibit smaller errors for relative elevation values close to 0. This is expected as most values in the dataset will lie within that range if the elevation patches were normalised (Figure \ref{fig:elevation_distributions}). With the elevation values diverging from the centre, predictions of both models get progressively worse, with the DepthAnythingV2's predictions deteriorating much more rapidly, apparent by the error distributions spreading significantly more, impacting its accuracy.\\

Despite the error difference, neither model is accurate enough for many of the DEM use cases. The mean average error is significantly higher than the horizontal DEM resolution, limiting its usability to potentially only big-scale geomorphological analyses. This indicates a potential for further studies focused on DEM generation through machine learning training models aiming to achieve higher accuracy, for example, through targeted fine-tuning, or direct elevation prediction in metric space. We see the \datasetname dataset as a valuable contribution to further research in this direction. 


\subsection{Limitations}
Despite significant effort put into the development of the \datasetname dataset, we want to highlight the following identified limitations:

\paragraph{\textbf{Limited dataset size}} the dataset provides 80,898 data samples, which would be considered a moderate size in the context of machine learning applications, where state-of-the-art solutions often employ datasets with millions \citep{depthv1}, sometimes billions of images \citep{laion5}, especially during pre-training phases. The small size of the dataset limits the size of models that can be trained using this dataset alone, due to the risk of over-parameterisation. This makes the dataset more suitable for fine-tuning scenarios or developing small, specialised models.

\paragraph{\textbf{Spatial bias}} despite best efforts put into eliminating biases in the dataset, some were unavoidable due to mentioned issues with the \citeauthor{day2023mars} repository, where a lot of invalid samples come from a specific longitudinal range $[-180^\circ, -120^\circ]$ or poor coverage of certain regions by the CTX instrument like the Hellas Planitia. Some features, such as the Olympus Mons, are absent in the dataset due to these reasons; therefore, we do not capture the entire longitudinal heterogeneity of Mars.

\paragraph{\textbf{Assumption of DEM and orthoimage alignment}} we do not conduct any sophisticated alignment efforts of the DEMs and orthoimages provided by the \citeauthor{day2023mars} repository. We process the samples with the assumption that the DEM and orthoimage can be superimposed directly after resizing. While we observed it to be true for a lot of the samples, we cannot guarantee the DEM and orthoimage patches are perfectly aligned.

\paragraph{\textbf{Imperfect outlier filtering}} the parameter selection process boils down to deriving a set of parameters that satisfy a trade-off of successful outlier identifications and minimising the false positive identifications. While we put our best efforts into picking a set of parameters with that trade-off in mind, numerous elevation outliers remain in the final dataset, and some useful features are flagged as outliers. This is why we made the binary masks available as a part of the dataset for its future users to decide how these locations should be handled.

\paragraph{\textbf{Relative depth evaluation}} we conduct the training and DEM generation in the standard score space, which corresponds to DEMs in relative depth space. We then use the ground-truth data to rescale the predictions into metric space, introducing a multiplicative error, which impacts accuracy in the metric space. For models aiming at higher prediction accuracy, DEM prediction directly in the metric space could be explored.

\section{Conclusions}
Our study aimed to address the critical need for high-quality training datasets in machine learning-based DEM generation. By creating the \datasetname dataset, we provide a valuable resource for developing and fine-tuning algorithms that can accurately predict DEMs from single optical images.\\

We view the DEM generation task as a derivative of the fundamental monocular depth estimation task; hence, we evaluate and compare the performance of a state-of-the-art DepthAnythingV2 foundation model with a significantly smaller baseline model trained on the \datasetname dataset for the DEM prediction task.\\

The results show the much smaller baseline model achieving a significantly better performance, indicating that MDE foundation models do not offer satisfactory zero-shot performance on DEM generation. These findings highlight the importance of the existence of the \datasetname dataset, to help develop new and fine-tune existing machine-learning algorithms, allowing for more accurate DEM prediction from single optical images.\\

The developed dataset makes creating machine learning solutions for DEM generation more accessible, providing high-quality, curated data suitable for model training. With the CTX instrument images having a coverage of nearly the entire surface of Mars, DEM generation solutions requiring only one perspective image would enable the generation of a new global Mars DEM with much higher resolution. The \datasetname dataset is a valuable addition in efforts heading towards that goal.

\section*{Acknowledgments}
We would like to acknowledge and express our gratitude to the creators of the \citeauthor{day2023mars} repository for generating high-quality data and making it openly available. This open-access nature of the data repository is invaluable for the scientific community to further analyse, process and develop new methodologies. Such contributions greatly enhance the accessibility of research; therefore, we are making the \datasetname dataset, together with all code necessary for its generation, fully open-source.

\bibliographystyle{abbrvnat}
\bibliography{main}

\appendix
\newpage
\section{Appendix} \label{appen}

\subsection{Processing parameters}
As described previously, the processing parameters have been chosen based on a manual adjustment process and visual inspection of the result in real time. We used 53 randomly selected patches coming from artefact-ridden samples from the \citeauthor{day2023mars} repository. Not all parameters were adjusted in this exact manner, as some of them contain just alternative methods, exchanging, for instance, bias for processing speed, and were used for development. Here we include the final set of parameters used in the dataset generation process. The complete set of all processing parameters can be found in the \texttt{patch\_generation/config/default\_config.py} file. Keep in mind that some parameters in this file are configuration parameters for methods that we experimented with and were not used in the processing pipeline in the end.

\begin{table}[!htb]
    \caption{Final dataset processing parameters}
    \centering
    \begin{tabular}{ll}
    \toprule
        Parameter name & Value \\
        \midrule
         \texttt{elevation\_map\_filling\_kernel} & 31 \\
          \midrule
         \texttt{local\_deviation\_sigma} & [1.2, 1.1, 0.9] \\
         \midrule
         \texttt{local\_deviation\_chunk\_size} & [10, 45, 90] \\
         \midrule
         \texttt{local\_deviation\_chunk\_overlap} & [5, 20, 30] \\
         \midrule
         \texttt{local\_deviation\_gaussian\_spread} & [0.2, 0.21, 0.27] \\
    \bottomrule
    \end{tabular}
    \label{tab:processing_parameters}
\end{table}

The samples used for patch generation for the parameter selection process were:
\begin{enumerate}
    \item \texttt{b03\_010628\_1974\_xn\_17n283w\_b01\_010206\_1975\_xn\_17n283w}
    \item \texttt{f09\_039203\_1984\_xn\_18n282w\_f09\_039348\_1984\_xn\_18n282w}
    \item \texttt{f13\_040970\_1974\_xn\_17n282w\_f13\_040904\_1974\_xn\_17n282w}
    \item \texttt{f16\_041893\_1974\_xn\_17n283w\_f14\_041115\_1974\_xn\_17n283w}
\end{enumerate}

\newpage
\subsection{Model architecture}
\begin{figure}[!htb]
    \centering
    \includegraphics[width=0.8\linewidth]{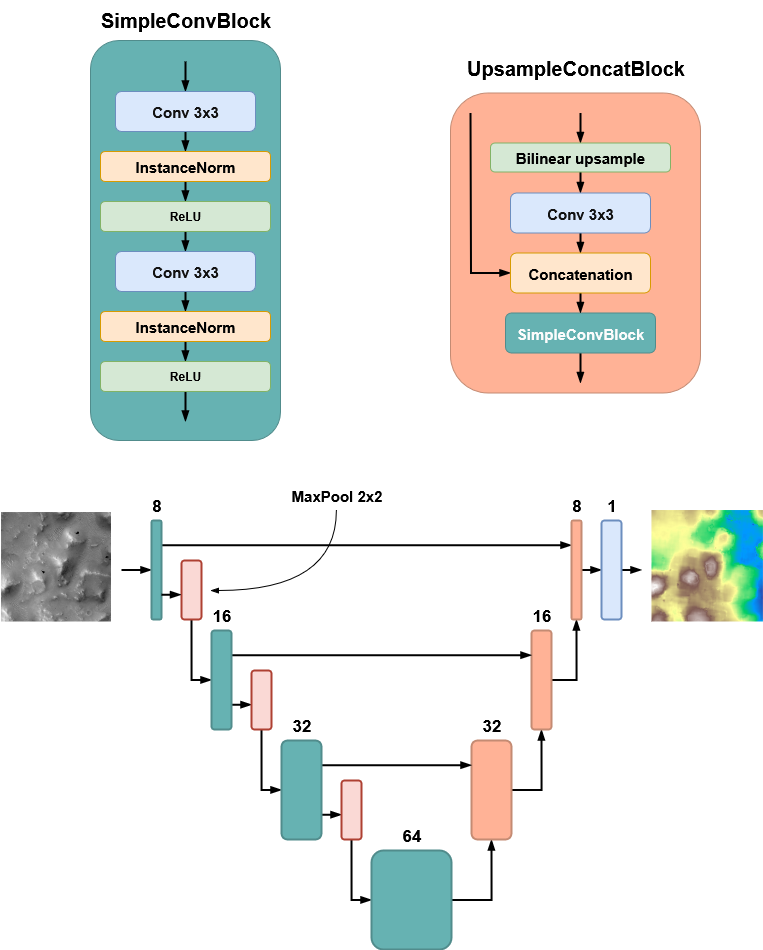}
    \caption{A diagram of the U-Net architecture used as a baseline performance model.}
    \label{fig:unet}
\end{figure}

\subsection{Verticalisation}
    The preprocessing starts by stripping away any spurious frame that is created when an oblique digital elevation map (DEM) is orthorectified.  
    Given an RGB image $I\in\mathbb{R}^{H\times W\times 3}$, rows (or columns) are iteratively discarded while the \emph{ratio} of black pixels in that line exceeds a user–defined parameter $\tau\in(0,1]$.  
    A first sweep uses the value $\tau=1.0$, meaning a row/column is rejected only when \emph{every} pixel is black.  
    
    To compute the image rotation needed to verticalise the image, two reference points are computed:  
    (i) the first non‑black pixel encountered while scanning down the leftmost column, giving its row coordinate $y_\ell$, and  
    (ii) the first non‑black pixel encountered while scanning rightwards along the bottom row, giving its column coordinate $x_b$.  
    Together with the fixed bottom‑left corner $(0,H-1)$ these points define a right‑angled triangle whose interior angle at $(0,H-1)$ is
    \begin{equation}
    \theta=\arctan\!\Bigl(\tfrac{H-y_\ell}{x_b}\Bigr),
    \qquad 
    \alpha=\tfrac{\pi}{2}-\theta .
     \end{equation}
    The image is \emph{clockwise} if $y_\ell<H/2$ and \emph{counter‑clockwise} otherwise; a horizontal mirror is applied in the former case so that rotation is always carried out in a consistent sense.  
    The image, the DEM and all auxiliary masks are then rotated by $-\alpha$ using a rigid Euclidean transform from
    \texttt{scikit-image} and \texttt{cv2} for any resampling.
    
    After rotation, a second sweep to eliminate black pixel rows/columns is executed with $\tau=0.1$, so a line is cropped when at least ten per cent of its pixels are black. This is usually aggressive enough to eliminate the thin wedges of near‑black padding introduced by interpolation, yet conservative enough to preserve genuine terrain in low‑contrast scenes.

    The end product is a verticalised, border‑free image and DEM.

\subsection{Deformed samples}

\begin{figure}[!htb]
   \centering
    \begin{subfigure}[b]{0.45\linewidth}
        \captionsetup{justification=centering}
        \includegraphics[width=\linewidth]{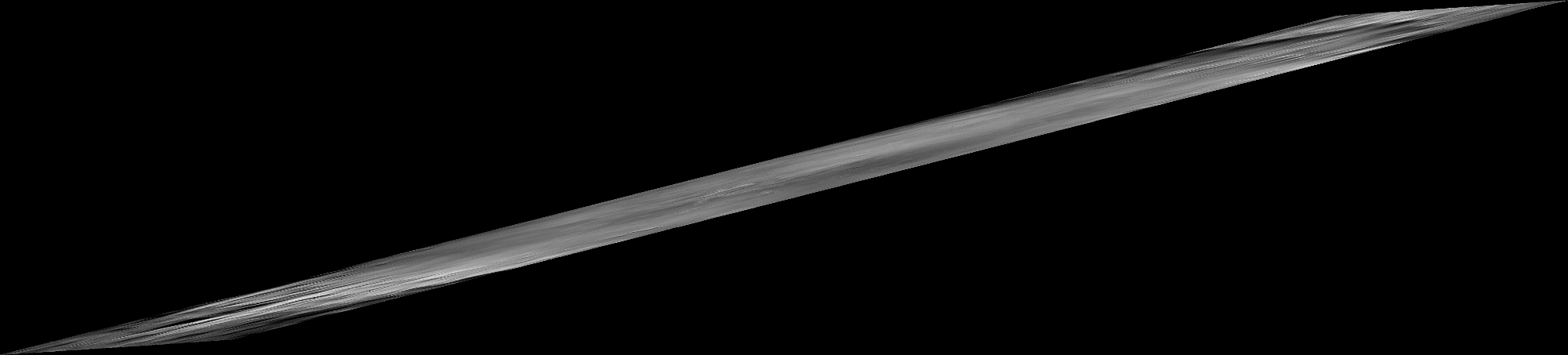}
        \caption[width=\linewidth]{\small \texttt{b04\_011310\_1396\_xi\_40s178w\_b06\_011811\_1396\_xn\_40s178w}\normalsize}
        \label{fig:distorted_0}
     \end{subfigure}
     \hfill
     \begin{subfigure}[b]{0.45\linewidth}
     \captionsetup{justification=centering}
        \includegraphics[width=\linewidth]{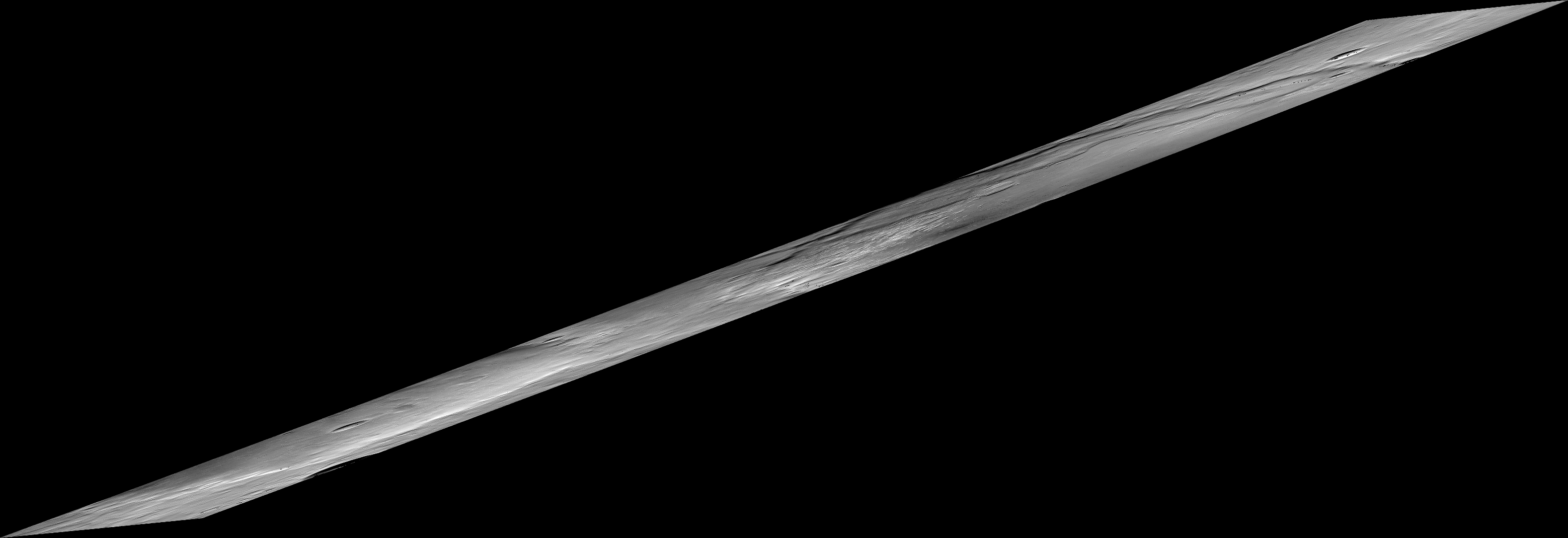}
        \caption{\small \texttt{b05\_011494\_1536\_xn\_26s160w\_b05\_011639\_1536\_xn\_26s160w}\normalsize}
        \label{fig:distorted_1}
     \end{subfigure}
     \vfill
     \begin{subfigure}[b]{0.5\linewidth}
     \captionsetup{justification=centering}
        \includegraphics[width=\linewidth]{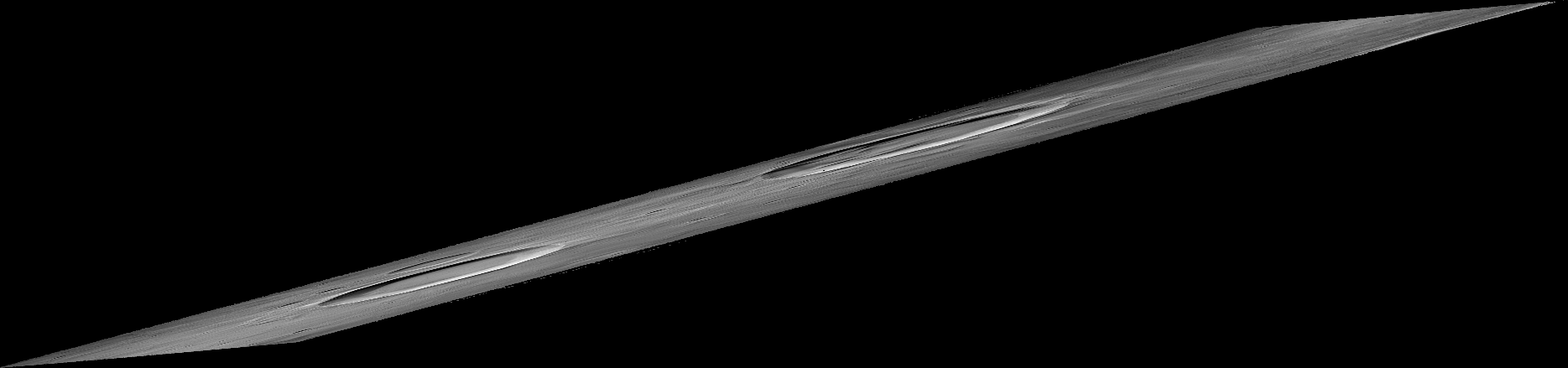}
        \caption{\small \texttt{b07\_012338\_1425\_xn\_37s161w\_b07\_012483\_1424\_xn\_37s161w}\normalsize}
        \label{fig:distorted_2}
     \end{subfigure}
     \caption{Examples of distorted samples in the \citeauthor{day2023mars} repository.}
     \label{fig:distorted_samples}
\end{figure}

\newpage
\subsection{Processing diagram}

\begin{figure}[!htb]
    \centering
    \includegraphics[width=0.8\linewidth]{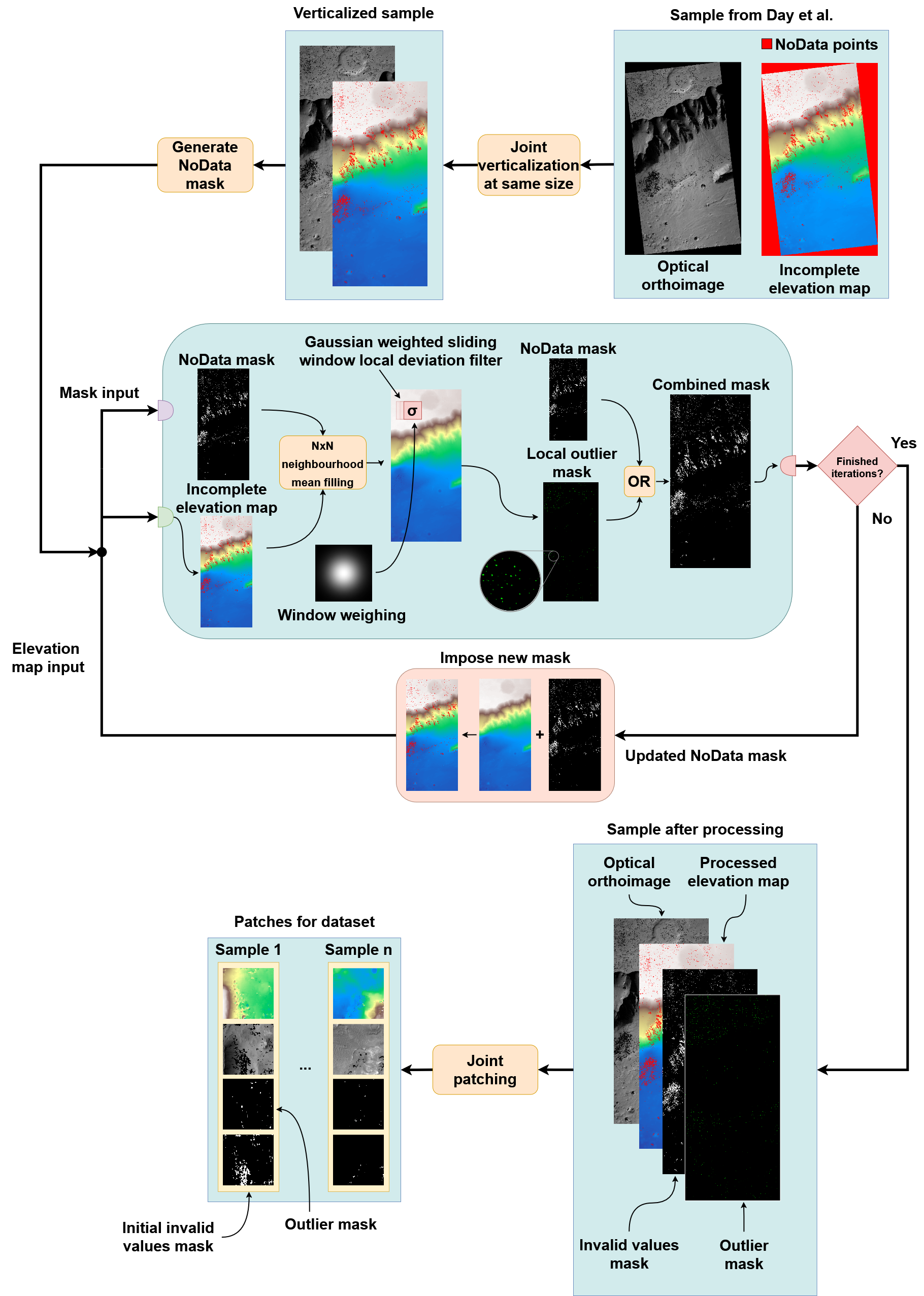}
    \caption{Schematic overview of the entire pipeline suited for processing samples from \citeauthor{day2023mars} into machine-learning-ready format.}
    \label{fig:processing_diagram}
\end{figure}

\end{document}